%% file: elsarticle-template-harv.tex
\documentclass[preprint,12pt,authoryear]{elsarticle}

\usepackage{amssymb}
\usepackage{amsmath}

\usepackage[T1]{fontenc}
\usepackage[utf8]{inputenc}
\usepackage{textcomp}
\usepackage{graphicx}
\usepackage{booktabs}
\usepackage{multirow}
\usepackage{xcolor}

\providecommand{\IEEEPARstart}[2]{\textbf{#1}#2}

\myfooterfont{\scriptsize\itshape}

\journal{ISPRS Journal of Photogrammetry and Remote Sensing}

\begin{document}

\begin{frontmatter}


\title{SpatialSR: Spatial Resolution-Centric Pipeline for Arbitrary-Scale Remote Sensing Super-Resolution} 

\author[aff1]{Yan Zhang}
\author[aff1]{Runfeng Luo}
\author[aff1]{Liangke Gao}
\author[aff1]{Lin Yuan}
\author[aff1]{Bin Xiao}
\author[aff1]{Xinbo Gao}

\affiliation[aff1]{organization={Chongqing University of Posts and Telecommunications},
            addressline={No. 2 Chongwen Road, Nan'an District},
            city={ChongQing},
            postcode={400065},
            country={China}}

\input{sec/0_abs}

\end{frontmatter}

\input{sec/1_intro}
\input{sec/2_related}
\input{sec/3_method}

\input{sec/4_experiment}

\input{sec/5_conclusion}

\appendix
\input{sec/6_sup}

\bibliographystyle{elsarticle-harv}
\bibliography{refs}

\end{document}

%% file: sec/0_abs.tex
\begin{abstract}
Although remote sensing image arbitrary-scale super-resolution (RSI-ASSR) methods support reconstruction at arbitrary or continuous scales, they predominantly condition on the relative scale factor and have not sufficiently investigated the absolute spatial resolution itself as a reconstruction prior, thereby disregarding the inherent spatial context of objects, which frequently leads to spatial-related distortions and compromised performance in downstream remote sensing applications.
To address these challenges, we introduce SpatialSR, a SRe-centric arbitrary-scale RSISR method that leverages the spatial resolution (SRe) of remote sensing images (RSI) as an essential prior for super-resolution reconstruction. 
Distinct from existing SR methods, SpatialSR aims to restore high-resolution remote sensing images with a target SRe (typically set to 0.1m/pixel to align with the metadata of commonly used downstream task datasets). 
Specifically, we design a Spatial Resolution Estimation Module (SREM) to estimate the SRe of low-resolution (LR) images, thereby dynamically determining the required arbitrary scaling factor. 
Next, we employ a pretrained SRe-to-image diffusion model to extract SRe-specified texture-style codes, which are further incorporated into a carefully designed Resolution-aware Style Consistency Module (RSCM) to constrain the SRe-aware texture consistency of reconstructed images. 
Furthermore, to overcome the high computational expense inherent in diffusion-based SR methods, we wrap the iterative diffusion process as a Trajectory Consistency (TC) loss, which enables comparable reconstructed texture quality of the generative model in SpatialSR within a single step. 
Extensive experiments demonstrate that the proposed SpatialSR achieves state-of-the-art (SOTA) performance on commonly used RSISR and downstream (image classification and segmentation tasks) datasets. 
\end{abstract}


%% file: sec/1_intro.tex
\section{Introduction}
\label{sec:intro}
\begin{figure}[t]
\centering
\includegraphics[trim=2px 43px 440px 8px, clip,width=\columnwidth]{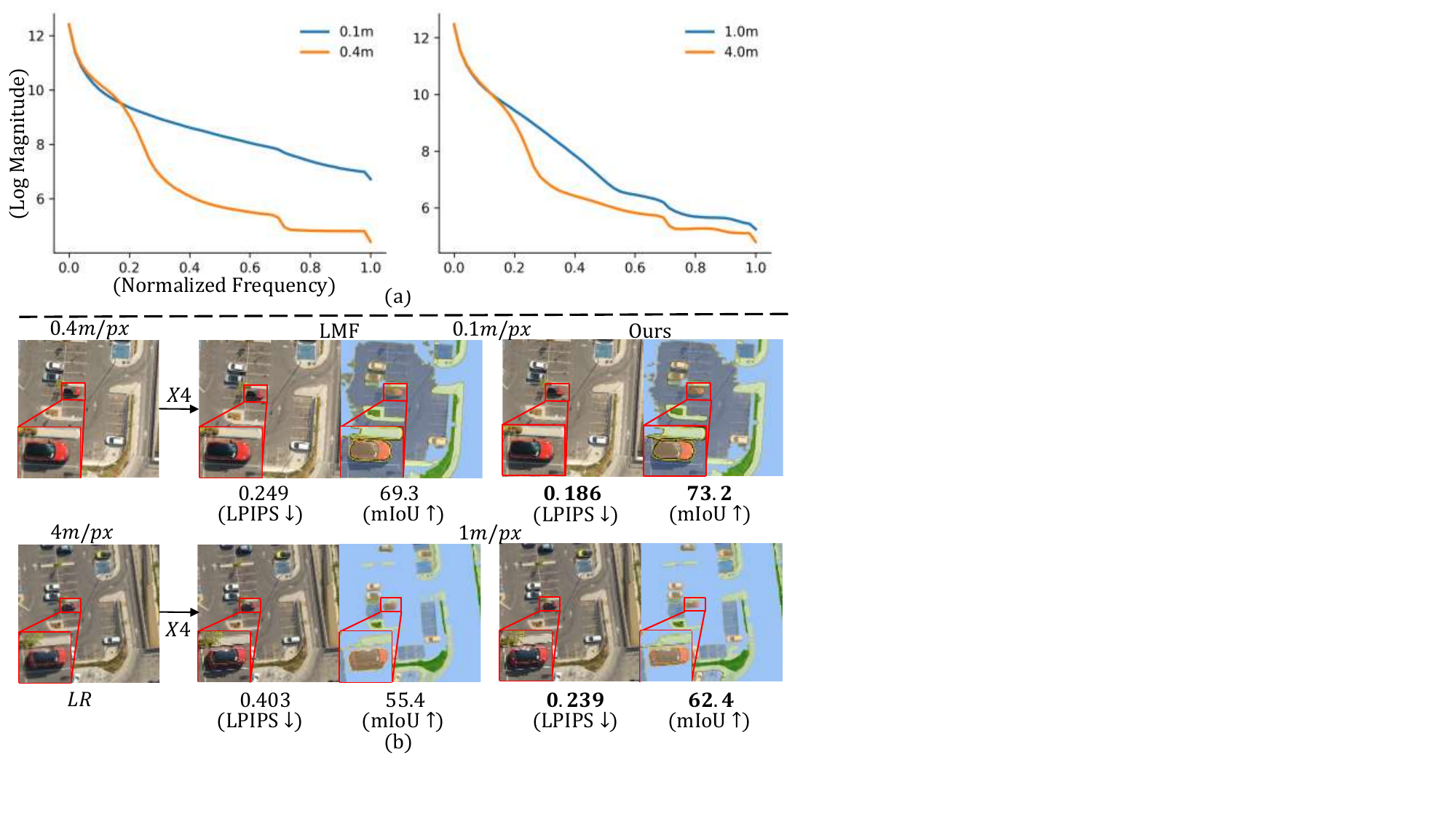}
\caption{Illustration of gradient statistics and downstream task performance at different spatial resolutions (SRe). (a) Comparison of average radial Fourier magnitude spectra under two spatial resolution settings. The left subplot compares images at 0.1 m/px with their 4× downsampled counterparts at 0.4 m/px, while the right subplot compares images at 1.0 m/px with their 4× downsampled counterparts at 4.0 m/px. For each SRe, the curve represents the mean radially averaged log-amplitude spectrum computed over all images, with the horizontal axis indicating normalized spatial frequency. (b) Visual and quantitative comparisons of ×4 super-resolution methods at different SRe.}
\label{fig:problem}
\end{figure} 

\IEEEPARstart{R}{emote} sensing image super-resolution (RSISR) aims to reconstruct high-resolution (HR) images from their low-resolution (LR) counterparts and serves as a fundamental preprocessing step for downstream remote sensing tasks such as semantic segmentation and scene classification. To break the fixed-scale limitation of conventional SR methods, arbitrary-scale super-resolution (ASSR) has attracted growing research interest: early approaches such as MetaSR~\cite{hu2019meta} dynamically predict upscale filter weights conditioned on the input scale, and LIIF~\cite{chen2021learning} further introduces implicit neural representations (INR) to model continuous image features, enabling rendering at arbitrary resolutions. These INR-based frameworks have been adapted to remote sensing (RS) and have achieved remarkable results through continuous-scale reconstruction~\cite{chen2023continuous} and scale-aware modulation~\cite{gao2024scale}.

However, existing remote sensing image arbitrary-scale super-resolution (RSI-ASSR) methods predominantly condition on the relative scale factor between the LR and HR images, while neglecting the absolute spatial resolution (SRe) itself as a reconstruction prior.
Consequently, these ASSR methods inevitably introduce significant texture ambiguity due to the ill-posed nature of this problem, where identical scaling factors can correspond to fundamentally different texture distributions. 
As illustrated in Fig.~\ref{fig:problem}, the texture distribution of HR images in the 4 m/pixel→1 m/pixel task differs substantially from that of the 0.4 m/pixel→0.1 m/pixel task, even though both of them are ×4 SR tasks. 
This discrepancy is also evident in the frequency domain, where distinct high-frequency energy distributions are observed across different resolutions, indicating significant cross-resolution distribution shifts. 
As a result, simply learning the LR-HR mapping without considering the specific spatial scale proves inadequate and tends to result in overfitting in the low-frequency region. 
This observation may partially explain why current SR methods often struggle in downstream applications, given that downstream performance heavily relies on high-frequency details. 
Compared with predefined scale factors, spatial resolution equivalence (SRe) serves as an essential scale metric for remote sensing imagery with inherent physical significance, as the morphological appearance of an object is highly correlated with its SRe. Throughout this paper, we distinguish the SRe --- an absolute physical attribute of a remote sensing image (measured in m/pixel and determined by the imaging altitude and sensor) --- from the scale factor, which is a relative ratio between the source and target SRe.
Based on the above analysis, an intuitive approach is to reconstruct SR images with specific SRe rather than predefined scale factors. 
However, in practice, the SRe of low-resolution (LR) imagery is frequently unavailable or unreliable due to missing metadata, unknown imaging conditions, or complex degradation processes. This limitation poses a fundamental obstacle to achieving high-quality reconstructed images with physical and spatial consistencies.

In addition to disregarding the physical SRe prior inherent to RSI, low-level vision tasks like RSISR are highly ill-posed, where one LR image has multiple HR candidates. 
Standard INR-based frameworks typically employ pixel-wise losses (e.g., L1 loss) for optimization, which are prone to generating over-smoothed textures.
To address this limitation, recent SR approaches have explored the adoption of generative models to supplement texture details. 
Compared with GAN~\cite{11237078} and VAE~\cite{10647786}, diffusion model~\cite{11094647} have demonstrated a remarkable capacity for synthesizing realistic high-frequency content within a harmonious theoretical framework. 
However, the iterative sampling process of diffusion models and their variants \cite{11243663} incurs substantial computational overhead and introduces significant inference latency, rendering them impractical for handling RSI with large resolution sizes. 

To address aforementioned challenges, we propose a novel RSISR method named SpatialSR, which is a flexible SRe-centric SR framework designed to synergistically combine physical prior-based texture reconstruction with enhanced adaptability to downstream SR tasks. 
SpatialSR is comprised of two core modules along with an explicitly designed Trajectory Consistency (TC) loss. 
The first module in SpatialSR is the SRe Estimation Module (SREM), which is designed to estimate the inherent SRe of LR remote sensing images and further dynamically calculate the precise upscaling factor with a predefined target SRe. 
Subsequently, we introduce a Resolution-aware Style Consistency Module (RSCM) to enhance the SRe-aware texture quality with diffusion prior from a pretrained SRe-to-image diffusion model. 
Finally, we wrap the diffusion process as a simple TC loss, which embeds the distribution modeling capability into the straight-forward SpatialSR model. 
The primary contributions of this study are summarized as follows.

\begin{itemize}
    \item To the best of our knowledge, this is the first study to explicitly investigate the incorporation of SRe into real-world RSI-ASSR tasks. The introduced SREM can not only estimate the source SRe from LR RS images but also extract SRe-aware feature representations.
    
    \item We introduce an RSCM that incorporates texture style priors derived from an pretrained diffusion model, enabling faithful and SRe-consistent texture reconstruction.

    \item We introduce a TC loss that wraps the diffusion process as a diffusion trajectory alignment constraint. With TC loss, the generative capability can be embedded into any conventional SR framework without iterative sampling.

\end{itemize}

%% file: sec/2_related.tex
\section{Related work}
\subsection{Arbitrary-Scale Super-Resolution}
ASSR techniques for natural images have been extensively explored. 
LIIF~\cite{chen2021learning} was the first to adapt the INR to the SR by using a multilayer perceptron to learn the mapping from image coordinates and features to RGB values. 
Subsequent studies have improved both the representational capacity and computational efficiency. 
LMF~\cite{He2024LatentMF} enhances local texture details by combining multi-frequency information in a computationally efficient manner, significantly reducing the computational costs while maintaining the reconstruction of fine-grained features. 
AnySR~\cite{Zhan2024AnySRRI} proposes a scale-adaptive unified framework that generalizes efficiently across upscaling factors. 
Recently, several studies have explored continuous modeling beyond traditional MLP-based INRs. GaussianSR~\cite{hu2025gaussiansr} represents images as continuous Gaussian fields, which improves the reconstruction of high-fidelity details. ContinuousSR~\cite{Peng2025PixelTG} formulates SR as a continuous rendering process, through which the model expressiveness can be further enhanced.

Despite these advances, the direct deployment of these methods on RS data remains challenging. Consequently, several studies have adapted continuous SR frameworks to RS scenarios.
For example, CSMNet~\cite{Liu2026CSMNetCM} proposes a channel-spatial Mamba UNet architecture for continuous remote sensing image super-resolution, through which long-range spatial dependencies can be effectively modeled.
BLiSR~\cite{wang2025binary} employs an adaptive binary continuous upsampling module that substantially reduces the computational and memory overhead. NeurOp-Diff~\cite{Xu2025NeurOpDiffCR} combines neural operators with diffusion priors for continuous-resolution remote sensing SR, whereas DISR~\cite{Liu2025DISRAN} performs dual-domain (spatial and frequency) implicit modeling for continuous remote sensing SR. 

Nevertheless, most ASSR methods are optimized with regression objectives and emphasize learning continuous mappings, which often yield over-smoothed reconstructed contexts and high-frequency noise. Moreover, remote sensing ASSR typically overlooks the performance of SR images in downstream tasks, which hinders the practical application of SR methods. Consequently, the current literature lacks both a consensus on these issues and a coherent, systematic framework to address them.

\subsection{Style consistency Modulation}
Style modulation and feature embedding are core techniques used in cross-domain RSI analysis. 
Early studies primarily relied on the Gram-matrix matching method proposed by Gatys et al.~\cite{Gatys2016ImageST}, through which style transfer was achieved by minimizing discrepancies in feature correlations.
Subsequently, feature modulation techniques such as AdaIN~\cite{Huang2017ArbitraryST} and FiLM~\cite{perez2018film} were introduced, which enabled the real-time processing of arbitrary style embeddings while maintaining a low computational cost.
StyleGAN~\cite{Karras2018ASG} maps random noise to style vectors, by which the attribute generation can be controlled across multiple scales.
SPADE~\cite{park2019semantic} further extends style injection from global to spatial dimensions, allowing precise spatial control.

In remote sensing, style modulation serves as an important tool for domain adaptation and multimodal fusion.
SGDM~\cite{wang2025semantic} further defines style as a disentangled imaging prior and effectively resolves texture-consistency issues in HR RS image synthesis and super-resolution by embedding semantic guidance or imaging-property constraints into the latent space of diffusion models. 
GeoSynth~\cite{sastry2024geosynth} integrates geolocation and text prompts to generate images with location-specific characteristics. 
STGAN~\cite{wang2025brain} performs style transfer using brain-inspired mechanisms, whereas VectorSynth~\cite{cher2026vectorsynth} maps vector-map attributes to semantic embeddings.

However, most existing style transfer methods heavily rely on predefined style images (e.g., same-sensor images or vector maps), which not only introduce additional data requirements and operational burden but also make the choice of style a direct determinant of the final performance. This creates an urgent need for style modulation approaches that do not rely on pre-defined style images.

\subsection{Diffusion-based Consistency Constraints}
Conventional diffusion-based image restoration typically relies on computationally expensive multi-step sampling, limiting its applicability to real-time remote sensing tasks. To address this, recent studies have recast diffusion priors as efficient consistency loss constraints. Consistency Models~\cite{Song2023ConsistencyM} first introduced a self-consistency constraint that maps any point along the diffusion trajectory directly to the initial state, thereby enabling single-step generation. Subsequently, DMD~\cite{yin2024improved} and the Phased Consistency Model~\cite{zheng2024trajectory} leverage distribution matching and phased trajectory alignment to enhance the texture of one-step outputs. In restoration, several studies~\cite{Wang2025SeedVR2OV} used pretrained diffusion models as score priors, enhancing the details by aligning the predicted restoration trajectories with the forward diffusion process. In SR-specific one-step generation, SinSR~\cite{Wang2023SinSRDI} distills a diffusion-based SR model into a single-step student, and CTMSR~\cite{You2025ConsistencyTM} explicitly matches the consistency trajectory between the one-step output and its multi-step teacher. Different from these methods, which distill a dedicated one-step SR network, our TC loss embeds the trajectory-alignment constraint directly into the training objective of any conventional SR backbone without altering its inference architecture. Closely related to our work, OmniScaleSR~\cite{Chai2025OmniScaleSRUS} injects a scale-controlled diffusion prior into ASSR to synthesize faithful and realistic textures; in contrast, SpatialSR conditions on the absolute SRe and distills the diffusion prior into a single-step TC loss without iterative sampling.

Despite reducing sampling complexity, many of these methods rely on complex teacher-student distillation schemes or require extensive fine-tuning of the diffusion model. Moreover, existing consistency losses often overlook the statistical observation model that links the restored outputs to the true data distribution.

%% file: sec/3_method.tex
\begin{figure}[t]
    \centering
    \includegraphics[width=\linewidth]{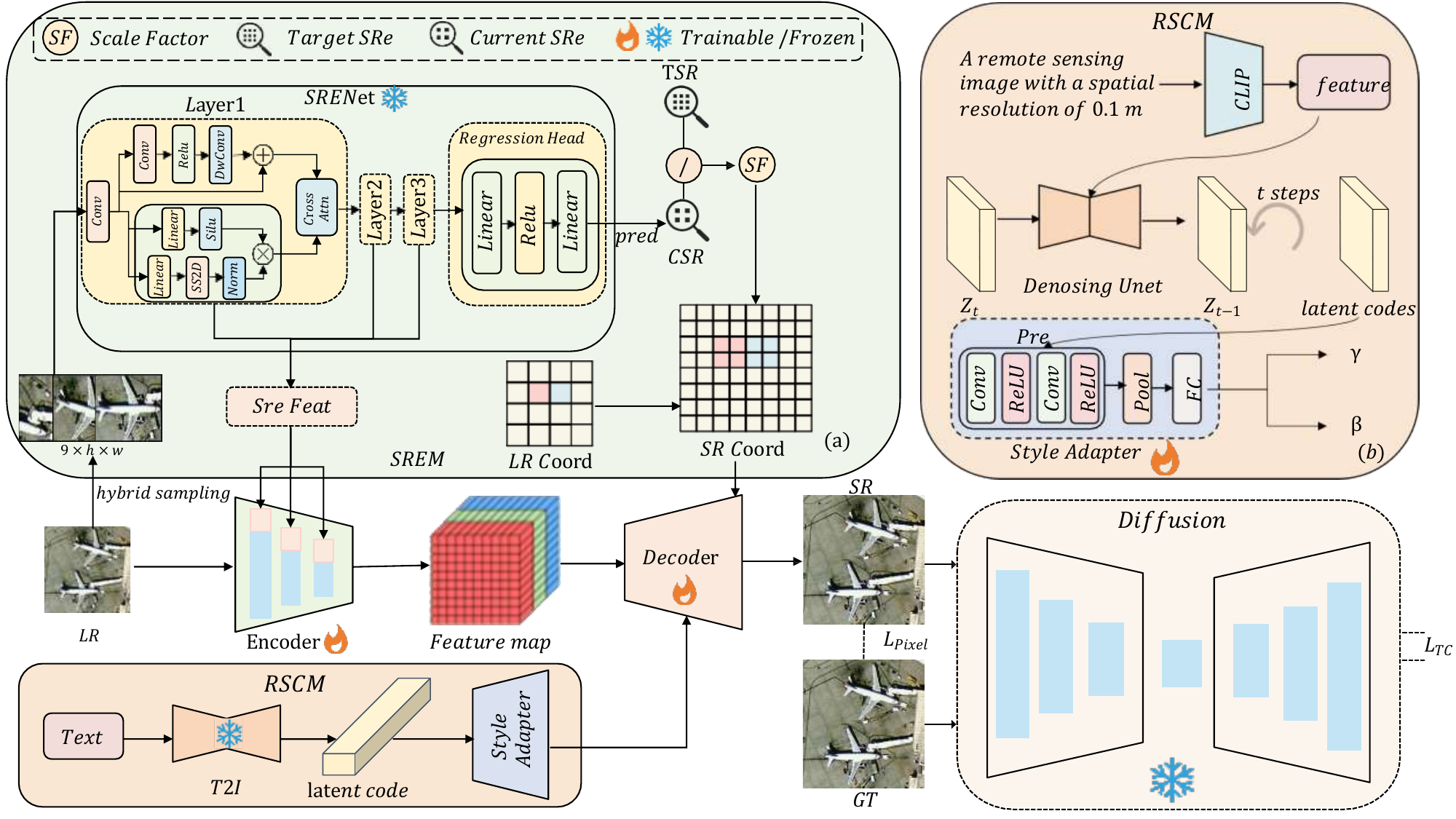}
    \caption{\textbf{Overview of the proposed SpatialSR pipeline.} Our pipeline consists of three core components. (a) The SREM estimates the source SRe from the LR image and then computes the scale factor together with the target SRe (Section~\ref{sec:srem_details}). (b) The RSCM synthesizes the final image under the modulation of a style embedding derived from the SRe-to-img model (Section~\ref{sec:decoder_details}). (c) Finally, the TC loss refines texture details by aligning the diffusion trajectories (Section~\ref{sec:dpr_details}).}
    \label{fig:overview}
\end{figure}

\section{Method}
\subsection{Overview}
As illustrated in Fig.~\ref{fig:overview}, the proposed SpatialSR is designed in an encoder-decoder fashion, comprising three key components: a Spatial Resolution Estimation Module (SREM), a Resolution-Aware Style Consistency Modulation (RSCM) module, and a Trajectory Consistency (TC) loss function. 
Specifically, given a low-resolution (LR) remote sensing image $I_{LR}$, we first feed it into the SREM to extract SRe-aware features $F_{sre}$ and estimate the corresponding spatial resolution equivalence $\hat{s}$. $F_{sre}$ is incorporated into the encoder to modulate the feature extraction process of the LR image.
Subsequently, the upscaling factor $\alpha$ is dynamically calculated by dividing the estimated SRe $\hat{s}$ by the target SRe.
Then, the RSCM module achieves SRe-aware texture priors $F_{style}$ via a pretrained generative SRe-to-image model. 
Finally, the modulated features, $\alpha$ and $F_{style}$ are passed into the decoder to obtain the reconstructed SR images.
During training, the SR image is optimized using the proposed TC loss to achieve superior perceptual performance. 

\subsection{Spatial Resolution Estimation and Scale Prediction}
\label{sec:srem_details}
SRe is a global metadata of RSI mainly determined by imaging altitude and sensor specifications. 
As a global attribute, SRe is naturally correlated with the frequency distribution of remote sensing imagery. 
However, accurately estimating SRe remains challenging due to the scarcity of frequency-aware clues in LR images and the inherent ill-posed nature of this inverse problem.

Based on the above analysis, we propose a concise yet effective hybrid feature sampling strategy and a hierarchical feature extractor termed SREM. 
First, we extract a set of candidate patches $\mathcal{P}$ using a sliding window mechanism with spatial size $p$ and stride $s$:
\begin{equation}
\mathcal{P} = \{ P_{i,j} \mid i \in \{0, s, \dots, W-p\}, j \in \{0, s, \dots, H-p\} \}, 
\label{sliding window}
\end{equation}
To identify regions that best represent specific image characteristics, we propose the top-k selection method to choose patches based on their patch-wise mean scores. Since SRe estimation relies on multiple aspects of image characteristics, we design three distinct operators $\{\mathcal{O}_1, \mathcal{O}_2, \mathcal{O}_3\}$.
\begin{align}
\mathcal{O}_{\text{1}}(i,j) &= 1, \label{random} \\
\mathcal{O}_{\text{2}}(i,j) &= \left(\nabla^2 G(i,j)\right)^2, \label{laplacian} \\
\mathcal{O}_{\text{3}}(i,j) &= \left(|LH| + |HL| + |HH|\right)_{\uparrow}(i,j). \label{haar}
\end{align}
For $\mathcal{O}_1$, we adopt a uniform random sampling strategy, motivated by the fact that remote sensing images are naturally dominated by low-frequency components. For $\mathcal{O}_2$, the Laplacian operator is used to evaluate high-frequency intensity, with top-30\% sampling to select regions with the highest average high-frequency response. For $\mathcal{O}_3$, the Haar wavelet transform is applied to capture and fuse multi-frequency structural representation.
Note that these patch-selection operators are non-differentiable and are applied as a data-preprocessing / inference-time selection step rather than as a learnable operation; therefore, they do not participate in back-propagation.

Subsequently, selected patches are concatenated along the channel dimension and fed into a carefully designed SRENet to predict the corresponding SRe sequence.
As illustrated in Fig.~\ref{fig:overview}(a), SRENet follows a hybrid-architecture paradigm, employing ResNet blocks~\cite{He2015DeepRL} at the first three layers to capture local features, followed by VisionMamba blocks~\cite{Zhu2024VisionME} to model global dependencies. An attention-based fusion module is designed to fuse and enhance cross-scale feature interactions.
With the estimated $\hat{s}$ and target SRe $s_{target}$, the upscaling factor $\alpha$ can be computed as:
\begin{equation}
\alpha = \frac{\hat{s}}{s_{target}}.
\label{eq:scale_factor}
\end{equation}

Finally, similar to INR methods~\cite{He2024LatentMF}, a continuous coordinate grid with the estimated $\alpha$ can be defined and fed to the next modules. To enhance feature representation, the spatial resolution features extracted by SREM are also fused with encoder outputs via concatenation and a simple convolutional block.
\begin{equation}
F_i^{\text{fused}}
=
\phi_i
\left(
\mathrm{Concat}
\left(
F_i^{\text{enc}},\;
F_i^{\text{srem}}
\right)
\right).
\label{fused}
\end{equation}

\subsection{Resolution-Aware Style Consistency Modulation}
\label{sec:decoder_details}
Since the texture distribution of remote sensing imagery varies across different SRe values, we introduce a Resolution-Aware Style Consistency Modulation (RSCM) into the SpatialSR decoder. 
First, we finetune the Stable Diffusion model~\cite{rombach2022high} on a SRe-to-image task to learn SRe-aware features, more details about training this model can be found in Section~\ref{sre-to-img}. 
With the pretrained model, we construct an SRe-aware prompt and an initial Gaussian noise vector as input. 
Following the standard denoising process of Stable Diffusion, we iteratively obtain the denoised image at timestep $t$. 
Because our prompt is semantic-irrelevant, the denoised image gradually contains clear SRe-aware style information without exact semantic meaning. 
We select the denoising latent feature at an earlier iteration as the extracted SRe-aware style embedding $F_{style}$:
\begin{equation}
F_{style} = \boldsymbol{x}_{t'} = \theta_{\text{early-stop}}(\boldsymbol{x}_T, c, t').
\label{eq:style_gen}
\end{equation}
where $t'$ is the stopped timestep, $F_{style}$ is the output of the VAE in $\theta$. To circumvent additional computational overhead during training, we extract $F_{style}$ offline. Specifically, we pre-compute the SRe-aware style information for target resolutions and construct an offline lookup dictionary, where each key represents a discrete SRe and the corresponding value denotes the pre-extracted latent code at that scale.
Since the dictionary keys are discrete, an intermediate SRe (e.g., $0.15$ m/px) is obtained by linear interpolation between its two nearest keys, i.e., $d_{0.15}=\alpha\,d_{0.1}+(1-\alpha)\,d_{0.2}$ with $\alpha\in[0,1]$. We acknowledge that a discrete dictionary cannot enumerate every continuous SRe value; the interpolation above provides a practical approximation, while the discrete-resolution assumption remains a limitation of our framework.

\begin{figure}[t]
    \centering
    \includegraphics[
        trim=0px 400px 540px 0px,
        clip,
        width=0.95\linewidth
    ]{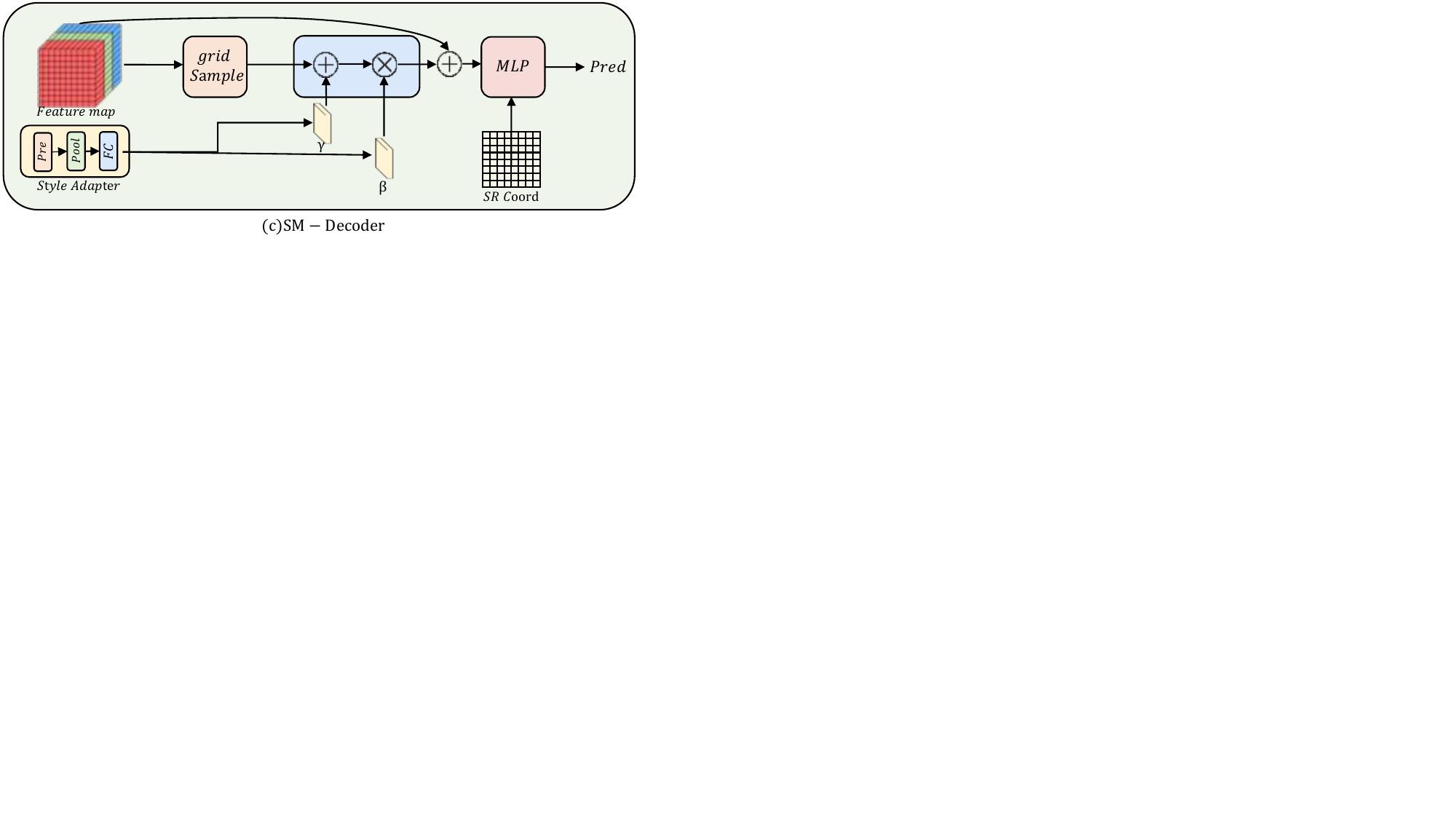}
    \caption{\textbf{Resolution-aware Style Consistency Modulation.} RSCM injects resolution-specific style information by modulating the decoder features. The style embedding is projected into scaling ($\boldsymbol{\gamma}$) and bias ($\boldsymbol{\beta}$) parameters, which are used to perform a residual affine transformation on the feature map.}
    \label{fig:sm_decoder}
\end{figure}

Next, we incorporate $F_{style}$ into the decoder via a FiLM-like mechanism~\cite{perez2018film}, as illustrated in Fig.~\ref{fig:sm_decoder}. In detail, the style embedding $F_{style}$ is firstly projected into a scale vector $\boldsymbol{\gamma}$ and a bias vector $\boldsymbol{\beta}$ via a lightweight adapter:
\begin{equation}
    \boldsymbol{\gamma}, \boldsymbol{\beta} = \text{Adapter}(F_{style}).
\end{equation}
These parameters then modulate a feature map $\mathbf{f}$ from the decoder via a residual affine transformation, where the tanh activation bounds the modulation factor:
\begin{equation}
    f_{\text{mod}} = f + (\tanh(\boldsymbol{\gamma}) \odot f + \boldsymbol{\beta}).
\label{eq:film}
\end{equation}
where $\odot$ denotes the element-wise product, and the $\tanh$ activation is employed to bound the modulation factor $\boldsymbol{\gamma}$.
As mentioned in~\cite{perez2018film}, this constraint prevents pulsed shifts of features and maintains training stability. The residual architecture enables the model to adaptively align feature statistics with the target SRe style while preserving structural integrity.

\begin{figure}[t]
    \centering
    \includegraphics[trim=0 247pt 0 25pt, clip, width=\linewidth]{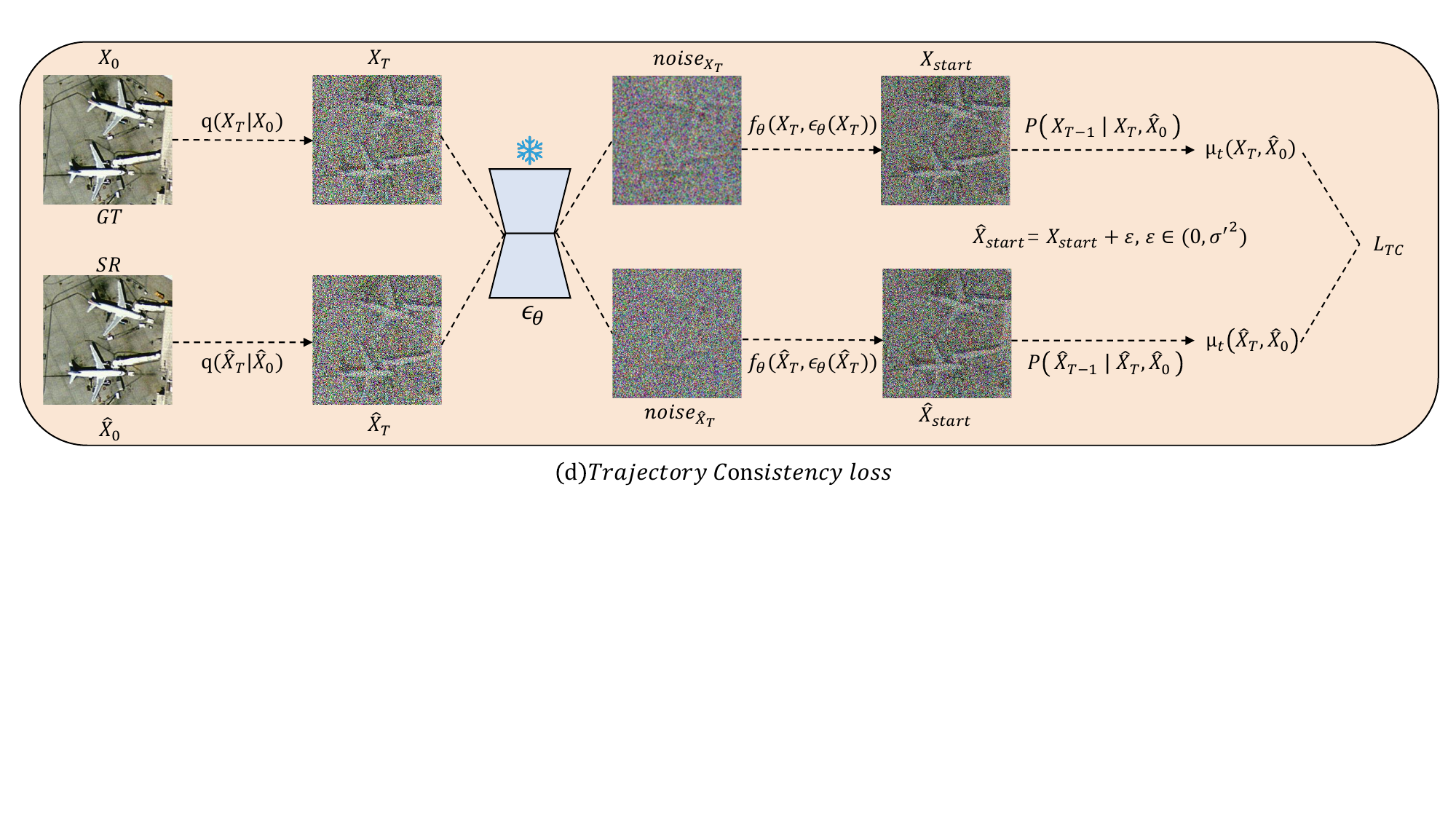}
    \caption{\textbf{Illustration of TC loss}. Both the SR image $\hat{\mathbf{X}}_0$ and its ground-truth (GT) counterpart $\mathbf{X}_0$ are passed through the denoising process of a pretrained diffusion model. The TC loss minimizes the discrepancy between their denoising trajectories, thereby encouraging the SR image to serve as a more realistic input to the diffusion model.}
    \label{fig:diffusion_loss}
\end{figure}
\subsection{Trajectory Consistency loss}
\label{sec:dpr_details}
To exploit the exceptional generative performance and texture synthesis capability of diffusion models, we introduce TC loss to incorporate high-fidelity textures into our reconstructed images. Unlike previous diffusion-based methods that require multi-step sampling, the proposed TC loss wraps the denoising process of the diffusion model into a trajectory alignment loss function. Firstly, we adopt the DDPM~\cite{Ho2020DenoisingDP} trajectory as our alignment target, since its explicitly defined Markovian diffusion process provides a well-structured and interpretable trajectory. Even though some of the latest mutations, such as DDIM~\cite{Song2020DenoisingDI}, Flowmatching~\cite{Lipman2022FlowMF}, achieve better inference efficiency, the reconstructed quality of DDPM is still satisfactory. As illustrated in Fig.~\ref{fig:diffusion_loss}, the key insight of TC loss is to explicitly map and calibrate the reconstructed HR images into the DDPM trajectory, thereby aligning the texture quality without multi-step denoising.

Given the clean latent variable $X_{0}$, its noisy counterpart $\hat{X}_{0}$, and standard Gaussian noise $\epsilon \sim \mathcal{N}(0,1)$, with timestep $T$ uniformly sampled from $\{0, \dots, 1000\}$, the insight of our TC loss is to calibrate the generative trajectory of the real data $X_T$ and its predicted counterpart $\hat{X}_T$. However, directly computing the distance between $P(X_{T-1}|X_{T})$ and $P(\hat{X}_{T-1}|\hat{X}_{T})$ is intractable due to their distinct conditioning variables. To address this issue, we reformulate the objective by conditioning both trajectories on the predicted clean latent $\hat{X}_0$:
\begin{equation}
\mathcal{L}_{\text{TC}} =\mathcal{D}\Big(P(X_{T-1}|X_{T},\hat{X}_{0})\;\mathbin{\|}\;P(\hat{X}_{T-1}|\hat{X}_{T},\hat{X}_{0})\Big),
\label{eq:main_objective}
\end{equation}
where $\mathcal{D}(\cdot)$ denotes a distance metric between the two distributions.
According to the standard DDPM formulation, the model-generated posterior 
$P(\hat{X}_{T-1}|\hat{X}_{T},\hat{X}_{0})$ 
follows a Gaussian distribution with analytical mean $\hat{\mu}$ and variance.
To derive the corresponding trajectory posterior $P(X_{T-1}|X_{T},\hat{X}_{0})$, we apply Bayes' theorem together with the Markovian property of the forward diffusion process, yielding:

\begin{equation}
P(X_{T-1}|X_{T},\hat{X}_{0})
=
\frac{
P(X_{T}|X_{T-1})
\, P(X_{T-1}|\hat{X}_{0})
}{
P(X_{T}|\hat{X}_{0})
},
\label{eq:bayes_theorem}
\end{equation}
where
\begin{equation}
P(X_{T}|X_{T-1})
=
\mathcal{N}
\big(
\sqrt{\alpha_T}X_{T-1},
\, 1-\alpha_T
\big).
\end{equation}
The remaining terms correspond to marginal likelihoods obtained by integrating out the true data $X_0$:

\begin{equation}
P(X_t|\hat{X}_0)
=
\int
P(X_t|X_0)
\, P(X_0|\hat{X}_0)
\, dX_0.
\label{eq:marginal}
\end{equation}
To derive a tractable closed-form solution for the integral in (\ref{eq:marginal}), we adopt a tractable local residual model that treats the reconstructed image $\hat{X}_0$ as a noisy observation of the clean image $X_0$:

\begin{equation}
\hat{X}_0 = X_0 + \epsilon', 
\qquad
\epsilon' \sim \mathcal{N}(0,\sigma'^2).
\end{equation}
Here the Gaussian residual form is a modeling assumption rather than a derived consequence of the observation model, adopted for tractability. Combined with the standard Gaussian prior for $X_0$, the posterior distribution $P(X_0|\hat{X}_0)$ remains Gaussian and is approximated as

\begin{equation}
P(X_0|\hat{X}_0)
\approx
\mathcal{N}(\hat{X}_0,\sigma'^2).
\end{equation}
Consequently, the marginal likelihood becomes Gaussian through convolution:

\begin{equation}
P(X_{T-1}|\hat{X}_0)
=
\mathcal{N}
\left(
\sqrt{\bar{\alpha}_{T-1}}\hat{X}_0,
\, 1-\bar{\alpha}_{T-1}
+ \bar{\alpha}_{T-1}\sigma'^2
\right).
\label{eq:posterior distribution}
\end{equation}

Similarly, the derivation of $P(X_T|\hat{X}_0)$ can be calculated with Eq.\ref{eq:posterior distribution}.
Putting these Gaussian marginals into (\ref{eq:bayes_theorem}),
the posterior distribution $P(X_{T-1}|X_T,\hat{X}_0)$ is also proven to be Gaussian:

\begin{equation}
P(X_{T-1}|X_T,\hat{X}_0)
=
\mathcal{N}(\mu,\sigma^2).
\end{equation}
Therefore, the analytical posterior mean $\mu$ can be computed as below
(full derivations are provided in the supplementary material):

\begin{equation}
\mu =
\frac{\sqrt{\bar{\alpha}_{T-1}}(1-\alpha_T)}
{1-\bar{\alpha}_T+\bar{\alpha}_T\sigma'^2}
\hat{X}_0
+
\frac{
(1-\bar{\alpha}_{T-1}+\bar{\alpha}_{T-1}\sigma'^2)
\sqrt{\alpha_T}
}{
1-\bar{\alpha}_T+\bar{\alpha}_T\sigma'^2
}
X_T .
\label{eq:posterior_mean}
\end{equation}
To simplify the probabilistic alignment in Eq.\ref{eq:main_objective}, we directly minimize the Mean Squared Error between the analytical posterior means, providing a tractable surrogate objective through mean alignment. With the above analysis, the final TC loss is defined as

\begin{equation}
\mathcal{L}_{\text{TC}}
=
\left\lVert
\mu - \hat{\mu}
\right\rVert_2^2,
\end{equation}

where

\begin{equation}
\hat{\mu}
=
\frac{\sqrt{\bar{\alpha}_{T-1}}(1-\alpha_T)}
{1-\bar{\alpha}_T}
\hat{X}_0
+
\frac{(1-\bar{\alpha}_{T-1})\sqrt{\alpha_T}}
{1-\bar{\alpha}_T}
\hat{X}_T.
\end{equation}

Finally, the overall loss function integrates the pixel-level $\mathcal{L}_1$ loss and the proposed TC loss:

\begin{equation}
\mathcal{L}
=
\mathcal{L}_{\text{pixel}}
+
\lambda
\mathcal{L}_{\text{TC}},
\label{eq:total_loss}
\end{equation}
where $\lambda$ balances the two losses, leading to a favorable perception-distortion trade-off.

%% file: sec/4_experiment.tex
\section{Experiments}

\begin{table}[t]
\centering
\caption{Quantitative comparison results of SR and downstream task performance on bicubic-degraded synthetic datasets.}
\setlength{\tabcolsep}{2.2pt}
\renewcommand{\arraystretch}{0.9}
\small
\resizebox{\textwidth}{!}{\begin{tabular}{c|c|cc|ccccc|ccccccccc}
\toprule
\multirow{2}{*}{Dataset} & \multirow{2}{*}{TSR} & \multirow{2}{*}{CSR} & \multirow{2}{*}{Metric}
& \multicolumn{5}{c}{Generative SR}
& \multicolumn{8}{c}{Implicit Neural Representation-based SR} \\
\cmidrule(r){5-9} \cmidrule(l){10-17}
& & & & ESRGAN & IDM & LDM & StableSR & ResShift & LIIF & ContSR & AnySR & SRNO & LMF & HIIF & Ours & LR \\
\midrule
\multirow{24}{*}{Vaihingen}
& \multirow{24}{*}{0.1}
& \multirow{6}{*}{0.5}
& PSNR↑ & 22.84 & 24.42 & 24.56 & 25.94 & 25.71 & 25.89 & 25.18 & 26.14 & 25.26 & \textbf{26.34} & 25.35 & \underline{26.21} &       &  \\
          &       &       & SSIM↑ & 0.6107 & 0.6873 & 0.6851 & 0.7589 & 0.7787 & 0.7795 & 0.7644 & 0.7841 & 0.7553 & \underline{0.7851} & 0.7699 & \textbf{0.7913} &       &  \\
          &       &       & LPIPS↓ & 0.3535 & 0.3583 & 0.3908 & \textbf{0.2479} & 0.2981 & 0.2687 & 0.2728 & 0.2744 & 0.2744 & 0.2746 & 0.2645 & \underline{0.2606} &       &  \\
          &       &       & DISTS↓ & 0.2011 & 0.2371 & 0.2606 & \textbf{0.1717} & 0.1936 & 0.1888 & 0.1898 & 0.1879 & 0.1879 & 0.1908 & 0.1942 & \underline{0.1875} &       &  \\
          &       &       & FID↓  & 99.78 & 96.26 & 119.89 & \textbf{57.03} & 58.86 & 67.18 & 69.23 & 64.95 & 76.74 & 66.23 & 73.80 & \underline{58.44} &       &  \\
          &       &       & mIoU↑ & 49.2  & 54.2  & 57.9  & 64.0    & 62.6  & 63.2  & 63.0    & \underline{64.3} & 63.8  & 64.2  & 63.6  & \textbf{66.7} & 55.7 \\
\cmidrule{3-17}
& & \multirow{6}{*}{1}
& PSNR↑ & 20.55 & 22.12 & 21.44 & 22.86 & 22.09 & 22.49 & 21.81 & 22.86 & 21.77 & \underline{23.04} & 21.90 & \textbf{23.33} &       &  \\
          &       &       & SSIM↑ & 0.5998 & 0.6472 & 0.6690 & 0.6773 & 0.6953 & 0.7045 & 0.6835 & 0.7103 & 0.6593 & \underline{0.7123} & 0.6849 & \textbf{0.7325} &       &  \\
          &       &       & LPIPS↓ & 0.5036 & 0.4759 & 0.4167 & \underline{0.3907} & \textbf{0.3656} & 0.4051 & 0.4185 & 0.4355 & 0.4137 & 0.4188 & 0.4044 & 0.4079 &       &  \\
          &       &       & DISTS↓ & 0.3269 & 0.3011 & 0.3273 & \textbf{0.2576} & 0.2653 & 0.2671 & 0.2709 & 0.2689 & 0.2657 & 0.2705 & 0.2668 & \underline{0.2602} &       &  \\
          &       &       & FID↓  & 181.39 & 171.29 & 208.88 & 145.03 & \textbf{131.43} & 142.79 & 155.46 & 142.62 & 159.47 & 146.08 & 150.50 & \underline{136.85} &       &  \\
          &       &       & mIoU↑ & 39.1  & 42.1  & 43.9  & 45.7  & \underline{46.6} & 41.3  & 43.2  & 43.1  & 42.2  & 43.9  & 42.0    & \textbf{50.7} & 38.2\\
\cmidrule{3-17}
& & \multirow{6}{*}{1.5}
& PSNR↑ & 20.57 & 20.51 & 20.53 & \underline{21.60} & 20.71 & 20.84 & 20.20 & 21.26 & 20.19 & 21.49 & 20.20 & \textbf{21.78} &       &  \\
          &       &       & SSIM↑ & 0.5948 & 0.6129 & 0.5867 & 0.6558 & 0.6387 & 0.6765 & 0.6555 & 0.6828 & 0.6325 & \underline{0.6858} & 0.6548 & \textbf{0.7117} &       &  \\
          &       &       & LPIPS↓ & 0.5362 & 0.5372 & 0.5276 & \textbf{0.4321} & \underline{0.4453} & 0.4917 & 0.5042 & 0.5281 & 0.5026 & 0.4961 & 0.4853 & 0.4638 &       &  \\
          &       &       & DISTS↓ & 0.3328 & 0.3491 & 0.3203 & \textbf{0.2753} & 0.2939 & 0.3096 & 0.3087 & 0.3102 & 0.2983 & 0.3106 & 0.2983 & \underline{0.2927} &       &  \\
          &       &       & FID↓  & 254.10 & 213.40 & 241.83 & 198.99 & \textbf{146.65} & 200.65 & 219.28 & 196.81 & 203.34 & 209.59 & 201.09 & \underline{193.43} &       &  \\
                    &       &       & mIoU↑ & 24.1  & 29.5  & 32.1  & 35.8  & \underline{36.4} & 31.7  & 32.4  & 29.7  & 30.3  & 32.5  & 34.2  & \textbf{40.3} & 21.5 \\
\cmidrule{3-17}
& & \multirow{6}{*}{2}
& PSNR↑ & 19.64 & 19.55 & 19.21 & 20.05 & 19.58 & 19.93 & 19.32 & 20.37 & 19.35 & \underline{20.62} & 19.27 & \textbf{21.02} &       &  \\
          &       &       & SSIM↑ & 0.5011 & 0.5948 & 0.5674 & 0.5906 & 0.5759 & 0.6644 & 0.6453 & 0.6719 & 0.6293 & \underline{0.6747} & 0.6436 & \textbf{0.7046} &       &  \\
          &       &       & LPIPS↓ & 0.5352 & 0.5653 & 0.5501 & 0.5029 & \textbf{0.4931} & 0.5481 & 0.5555 & 0.5837 & 0.5497 & 0.5418 & 0.5324 & \underline{0.5011} &       &  \\
          &       &       & DISTS↓ & 0.3352 & 0.3787 & 0.3296 & 0.3516 & 0.3679 & 0.3386 & 0.3368 & 0.3411 & 0.3219 & 0.3397 & \underline{0.3191} & \textbf{0.3148} &       &  \\
          &       &       & FID↓  & 286.52 & 239.79 & 233.60 & \textbf{227.61} & 233.19 & 252.95 & 272.91 & 243.68 & 241.74 & 254.77 & 239.65 & \underline{230.99} &       &  \\
          &       &       & mIoU↑ & 24.9  & 28.7  & 30.6  & 29.5  & \underline{30.7} & 27.9  & 26.8  & 24.8  & 28.1    & 29.4  & 30.0    & \textbf{31.5} & 11.7 \\
\midrule
\multirow{24}{*}{UCMD}
& \multirow{24}{*}{0.5}
& \multirow{6}{*}{2}
& PSNR↑ & 24.30 & 24.57 & 24.74 & 24.78 & 25.87 & 28.88 & 28.91 & 28.84 & 28.99 & 28.71 & \underline{29.01} & \textbf{29.12} \\
          &       &       & SSIM↑ & 0.6294 & 0.5974 & 0.6592 & 0.6063 & 0.6850 & 0.7834 & 0.7841 & 0.7814 & \underline{0.7885} & 0.7794 & 0.7875 & \textbf{0.8036} \\
          &       &       & LPIPS↓ & 0.2698 & 0.2721 & 0.2561 & 0.2396 & \textbf{0.2264} & 0.2846 & 0.2732 & 0.2812 & 0.2622 & 0.2848 & 0.2762 & \underline{0.2359} \\
          &       &       & DISTS↓ & 0.2456 & 0.2161 & 0.2095 & \textbf{0.2055} & 0.2098 & 0.2174 & 0.2173 & 0.2125 & 0.2145 & 0.2166 & 0.2147 & \underline{0.2075} \\
          &       &       & FID↓  & 89.14 & 121.78 & 79.01 & 91.80& \textbf{69.79} & 90.82 & 91.42 & 83.83 & 88.36 & 86.15 & 88.36 & \underline{76.57} \\
          &       &       & Acc↑ & 96.81 & 88.29 & 97.08 & \underline{97.39} & 97.01    & 97.10  & 97.33 & 97.24 & 97.26 & 97.05 & 97.38 & \textbf{97.48} & 94.92 \\
\cmidrule{3-17}
& & \multirow{6}{*}{4}
& PSNR↑ & 23.02 & 22.70 & 22.77 & 23.05 & 22.49 & 24.87 & 24.94 & 24.89 & 24.99 & 24.83 & \underline{25.02} & \textbf{25.29} \\
          &       &       & SSIM↑ & 0.5175 & 0.5346 & 0.5917 & 0.5011 & 0.4883 & 0.6068 & 0.6083 & 0.6041 & 0.6125 & 0.6017 & \underline{0.6128} & \textbf{0.6309} \\
          &       &       & LPIPS↓ & 0.2991 & \textbf{0.2871} & \underline{0.2921} & 0.3213 & 0.4206 & 0.3168 & 0.3175 & 0.3158 & 0.3129 & 0.3207 & 0.3126 & 0.2929 \\
          &       &       & DISTS↓ & 0.2891 & \textbf{0.2871} & 0.2921 & 0.3213 & 0.2973 & 0.3168 & 0.3175 & 0.3158 & 0.3129 & 0.3207 & 0.3126 & \underline{0.2882} \\
          &       &       & FID↓  & 116.55 & 148.69 & 113.97 & 116.03 & \textbf{101.64} & 155.31 & 151.65 & 140.45 & 145.45 & 142.54 & 146.90 & \underline{111.44} \\
          &       &       & Acc↑ & 85.11 & 77.14 & 89.74 & 83.57 & 89.29 & 90.38 & 90.71 & 90.43 & 90.93  & 90.33 & \underline{92.14} & \textbf{93.24} & 70.24 \\
\cmidrule{3-17}
& & \multirow{6}{*}{6}
& PSNR↑ & 19.43 & 20.63 & 20.43 & 22.22 & 21.16 & 22.62 & 22.64 & \underline{22.73} & 22.44 & 22.72 & 22.59 & \textbf{23.15} \\
          &       &       & SSIM↑ & 0.4579 & 0.4359 & 0.4229 & 0.4705 & 0.4126 & 0.5201 & 0.5221 & 0.5197 & 0.5192 & 0.5176 & \underline{0.5232} & \textbf{0.5946} \\
          &       &       & LPIPS↓ & 0.5857 & 0.6051 & 0.5508 & \textbf{0.5259} & 0.5541 & 0.6432 & 0.6519 & 0.6779 & 0.6181 & 0.6917 & 0.6143 & \underline{0.5483} \\
          &       &       & DISTS↓ & 0.3534 & 0.3862 & 0.3396 & 0.3437 & \underline{0.3332} & 0.3717 & 0.3704 & 0.3718 & 0.3614 & 0.3834 & 0.3575 & \textbf{0.3312} \\
          &       &       & FID↓  & 151.79 & 182.65 & \textbf{132.67} & 142.85 & 143.37 & 187.66 & 181.54 & 169.60& 172.81 & 180.47 & 172.10 & \underline{136.33} \\
          &       &       & Acc↑ & 64.57 & 67.24 & 71.00    & 74.59 & 76.57 & 71.76 & 73.97  & 72.38 & 77.71 & 68.52 & \underline{78.14} & \textbf{80.86} & 33.19 \\
\cmidrule{3-17}
& & \multirow{6}{*}{8}
& PSNR↑ & 19.68 & 20.41 & 20.56 & 21.56 & 20.82 & 21.80 & 21.85 & \underline{21.91} & 21.62 & 21.90 & 21.76 & \textbf{22.09} \\
          &       &       & SSIM↑ & 0.4201 & 0.4005 & 0.4272 & 0.4536 & 0.4082 & 0.4889 & \underline{0.4903} & 0.4885 & 0.4862 & 0.4868 & 0.4897 & \textbf{0.5028} \\
          &       &       & LPIPS↓ & 0.5669 & 0.5976 & 0.5848 & \underline{0.5271} & 0.6014 & 0.7116 & 0.7293 & 0.7541 & 0.6998 & 0.7627 & 0.6894 & \textbf{0.5258} \\
          &       &       & DISTS↓ & 0.3869 & 0.3931 & 0.3618 & \underline{0.3592} & 0.3682 & 0.4096 & 0.4102 & 0.4112 & 0.3949 & 0.4237 & 0.3902 & \textbf{0.3355} \\
          &       &       & FID↓  & 177.41 & 189.25 & \underline{150.18} & 167.15 & 162.90 & 215.10 & 207.07 & 197.58 & 200.04 & 212.06 & 195.53 & \textbf{138.02} \\
          &       &       & Acc↑ & 47.62 & 46.82 & 58.43 & 57.71 & \underline{59.64} & 50.05 & 53.05 & 49.76 & 57.76 & 45.76 & 58.86 & \textbf{62.71} & 14.95 \\
\bottomrule
\end{tabular}}

\label{tab:0.1tgt}
\end{table}

\subsection{Experimental Setup}
\label{sre-to-img}
In all tables, bold numbers indicate the best result and underlined numbers indicate the second-best result.

\textit{1)} Datasets:
To evaluate the image reconstruction quality and downstream task performance of SpatialSR under both synthetic and real-world scenarios, experiments are conducted on three semantic segmentation benchmarks, including ISPRS-Vaihingen, LoveDA-val, and UAVid, as well as two image classification datasets, namely UCMD and WHU-RS19. In addition, we introduce a Multi-Spatial Resolution Unmanned Aerial Vehicle (MSRUAV) dataset to support model training and further validate the robustness of the proposed method. In the synthetic scenario, original images are resampled to target spatial resolutions as ground-truth (GT) references. SpatialSR reconstructs the corresponding LR inputs to the target resolutions, and the resulting performance gains in downstream tasks are compared with LR-based results. In the real-world scenario, source images are directly processed by the SR model to reach specified target resolutions without additional degradation.

 
\textit{2)} Training Details:
Arbitrary-scale LR-HR pairs are generated using the Real-ESRGAN degradation pipeline and bicubic downsampling. Resolution-aware latent priors extracted from an SRe-to-image diffusion model are incorporated into RSCM to guide spatial-resolution-aware reconstruction. Additional implementation details are provided in the supplementary material.

\textit{3)} Metrics:
We evaluate reconstruction quality using PSNR, SSIM, LPIPS, DISTS, and FID, covering both fidelity and perceptual aspects. To further assess no-reference image quality, NIQE, MUSIQ, MANIQA, and CLIPIQA are also reported. In addition, downstream task performance is evaluated using mIoU for semantic segmentation and Top-1 Accuracy for image classification. All downstream models are applied to the SR images in a zero-shot manner using their officially released weights without additional fine-tuning: semantic segmentation uses UNetFormer from the GeoSeg framework, and image classification uses ResNet-18.

\textit{4)} Comparison Methods:
We compare SpatialSR with representative generative-based methods, including ESRGAN~\cite{wang2018esrgan}, IDM~\cite{gao2023implicit}, LDM~\cite{Rombach2021HighResolutionIS}, StableSR~\cite{Wang2023ExploitingDP}, and ResShift~\cite{Yue2023ResShiftED}, as well as INR-based methods, including LIIF~\cite{Chen2020LearningCI}, ContinuousSR (ContSR)~\cite{Peng2025PixelTG}, AnySR~\cite{Zhan2024AnySRRI}, SRNO~\cite{Wei2023SuperResolutionNO}, LMF~\cite{He2024LatentMF}, and HIIF~\cite{Jiang2024HIIFHE}. To ensure a fair evaluation under the ASSR setting, all fixed-scale SR baselines are further fine-tuned, and the better-performing variant of each method is reported. The lightweight INR-based method LMF also serves as the shared backbone baseline in our ablation study.

\subsection{Quantitative Evaluation}
Performance comparisons of super-resolution and downstream tasks across both synthetic and real-world scenarios, where synthetic datasets contain paired ground truth references that are generated through controlled degradation processes while real-world datasets are directly collected without corresponding high resolution references, are summarized in Tables~\ref{tab:0.1tgt} and \ref{tab:true}.
Performance analysis with metrics of fidelity, referenced perceptual quality, non referenced perceptual quality, and downstream tasks, which are adopted according to the availability of ground truth references in different evaluation scenarios, is elaborated below:

\begin{table}[t]
\centering
\caption{Quantitative comparison results of remote-sensing image SR under real-world scenarios.}
\setlength{\tabcolsep}{1.8pt}
\renewcommand{\arraystretch}{0.9}
\small
\resizebox{\textwidth}{!}{
\begin{tabular}{c|cc|c|ccccc|ccccccccc}
\toprule
\multirow{2}{*}{Dataset} & \multirow{2}{*}{TSR} & \multirow{2}{*}{CSR} & \multirow{2}{*}{Metric} 
& \multicolumn{5}{c}{Generative SR} 
& \multicolumn{8}{c}{Implicit Neural Representation-based SR} \\
\cmidrule(r){5-9} \cmidrule(l){10-17}
& & & & ESRGAN & IDM & LDM & StableSR & ResShift & LIIF & ContSR & AnySR & SRNO & LMF & HIIF & Ours & Ori \\
\midrule
\multirow{5}{*}{UAVid}
& \multirow{5}{*}{0.1}
& \multirow{5}{*}{0.16}
& NIQE↓ & 5.11 & 6.10 & 4.99 & 5.34 & \textbf{4.74} & 5.54 & 5.22 & 5.53 & 5.39 & 5.16 & 5.44 & \underline{4.92} &  \\
          &       &       & MUSIQ↑ & 60.6 & 52.6 & 51.8 & 63.0 & \underline{64.4} & 59.4 & 60.9 & 58.6 & 60.6 & 60.0 & 60.6 & \textbf{64.7} &  \\
          &       &       & MANIQA↑ & 0.42 & 0.39 & 0.53 & 0.61 & 0.58 & 0.64 & 0.66 & 0.67 & \underline{0.68} & 0.65 & 0.64 & \textbf{0.69} &  \\
          &       &       & CLIPIQA↑ & \underline{0.50} & 0.28 & 0.28 & 0.47 & \textbf{0.54} & 0.38 & 0.36 & 0.39 & 0.37 & 0.35 & 0.37 & 0.48 &  \\
          &       &       & mIoU↑ & 72.2 & 71.7 & 72.2 & 74.6 & \underline{74.7} & 73.4 & 73.5 & 73.4 & 73.4 & 73.6 & 73.4 & \textbf{75.8} & 72.5 \\
\cmidrule{1-17}
\multirow{5}{*}{WHU-RS19}
& \multirow{5}{*}{0.1}
& \multirow{5}{*}{0.5}
& NIQE↓ & 6.65 & 6.87 & 6.47 & 6.99 & \underline{6.45} & 7.83 & 8.36 & 8.48 & 9.92 & 7.58 & 7.90 & \textbf{5.90} &  \\
          &       &       & MUSIQ↑ & 42.9 & 39.1 & 41.8 & \underline{45.2} & 43.4 & 35.3 & 36.0& 31.1 & 30.5 & 37.3 & 35.0 & \textbf{46.5} &  \\
          &       &       & MANIQA↑ & 0.37 & 0.35 & 0.47 & \textbf{0.49} & 0.45 & 0.32 & 0.32 & 0.36 & 0.43 & 0.32 & 0.33 & \underline{0.48} &  \\
          &       &       & CLIPIQA↑ & 0.25 & 0.29 & 0.28 & \textbf{0.41} & \underline{0.40} & 0.12 & 0.32 & 0.13 & 0.27 & 0.13 & 0.12 & 0.31 &  \\
          &       &       & Acc↑ & 96.5 & 96.6 & 96.6 & \underline{96.8} & 96.6 & 96.5 & 96.6 & 96.5 & 96.6 & 96.5 & 96.7 & \textbf{97.9} & 95.8 \\
\bottomrule
\end{tabular}}
\label{tab:true}
\end{table}

\textbf{Fidelity}. 
Table~\ref{tab:0.1tgt} illustrates that INR-based methods such as LMF and HIIF consistently yield higher PSNR and SSIM across various SR scales rather than generative methods. 
However, this capacity is gradually diminished at larger upscaling factors, due to pronounced over-fitting on smoothing area. 
This phenomenon is widely observed and analyzed in generative SR approaches~\cite{Ledig2016PhotoRealisticSI}, which indicate that methods optimized toward pixel level reconstruction objectives generally achieve higher PSNR values while producing over-smoothed results with insufficient texture details and perceptually unsatisfactory appearances.
Conversely, generative models typically exhibit lower fidelity due to hallucinations that deviate from the GT geometry. 
Among generative methods, the proposed SpatialSR maintains superior PSNR/SSIM metrics, which indicates the critical effectiveness of SRe prior in reconstructing high-fidelity textures.

\textbf{Perceptual quality}. 
On perceptual metrics (LPIPS, DISTS, FID), generative approaches generally outperform INR-based methods with a large gap by reconstructing visually richer textures. 
Nevertheless, methods such as ResShift may achieve favorable LPIPS/DISTS scores by introducing excessive high-frequency artifacts and therefore degrade referenced perceptual metrics like FID. 
In contrast, the robust performance of SpatialSR across all perceptual metrics validates its ability to recover realistic textures with the SRe-aware style consistency.
Notably, for high SR scales such as ×8 on UCMD, where ResShift and StableSR exhibit clear degradation in perceptual fidelity and distribution consistency, SpatialSR still preserves both qualities. It surpasses ResShift by 12.6\% in LPIPS and StableSR by 17.4\% in FID, which further demonstrates the superiority of SpatialSR under challenging SR scenarios.

The no-referenced perceptual metrics (e.g., NIQE, MANIQA) are illustrated in Table~\ref{tab:true}. 
Generative methods show advantages in blind quality assessment, whereas INR-based methods lag behind due to overly smooth reconstructions. 
The robustness of SpatialSR in practical applications is evidenced by its superior performance across these metrics, even under challenging conditions such as the 5× real-world degradation on the WHU-RS19 dataset. 
This suggests that our unified modeling of structure and texture generalizes well beyond synthetic settings.

\textbf{Downstream tasks}. 
An interesting result is observed that neither INR-based methods nor generative methods achieve satisfactory performance on semantic segmentation and image classification. 
INR-based methods tend to miss small objects due to insufficient texture details, while generative methods may introduce confused semantic artifacts. 
Counterintuitively, there is less correlation observed between perceptual or fidelity metrics and downstream performance.
Specifically, as shown in Table~\ref{tab:0.1tgt}, generative models like ResShift often achieve the best LPIPS or FID scores (e.g., a leading LPIPS of 0.3656 in Vaihingen at x10), yet their mIoU lags significantly behind, falling over 10\% lower than our method.
The superior mIoU and accuracy achieved by SpatialSR testify to its capacity for recovering semantically meaningful textures while maintaining structural integrity. 
For example, on the Vaihingen dataset, our method surpasses the current state-of-the-art (ResShift) by 4.1 in the ×10 scenario, representing a substantial improvement. Similarly, in terms of Acc on the UCMD dataset, our method achieves further gains in the ×12 scenario, surpassing the current state-of-the-art (HIIF) by 2.72.
The strong correlation between SR quality and downstream performance further confirms that our design effectively achieves target SRe reconstruction with SREM.

\subsection{Qualitative Comparison}

\begin{figure}[t]
    \centering
    \includegraphics[trim=5px 112px 10px 0, clip,width=\linewidth]{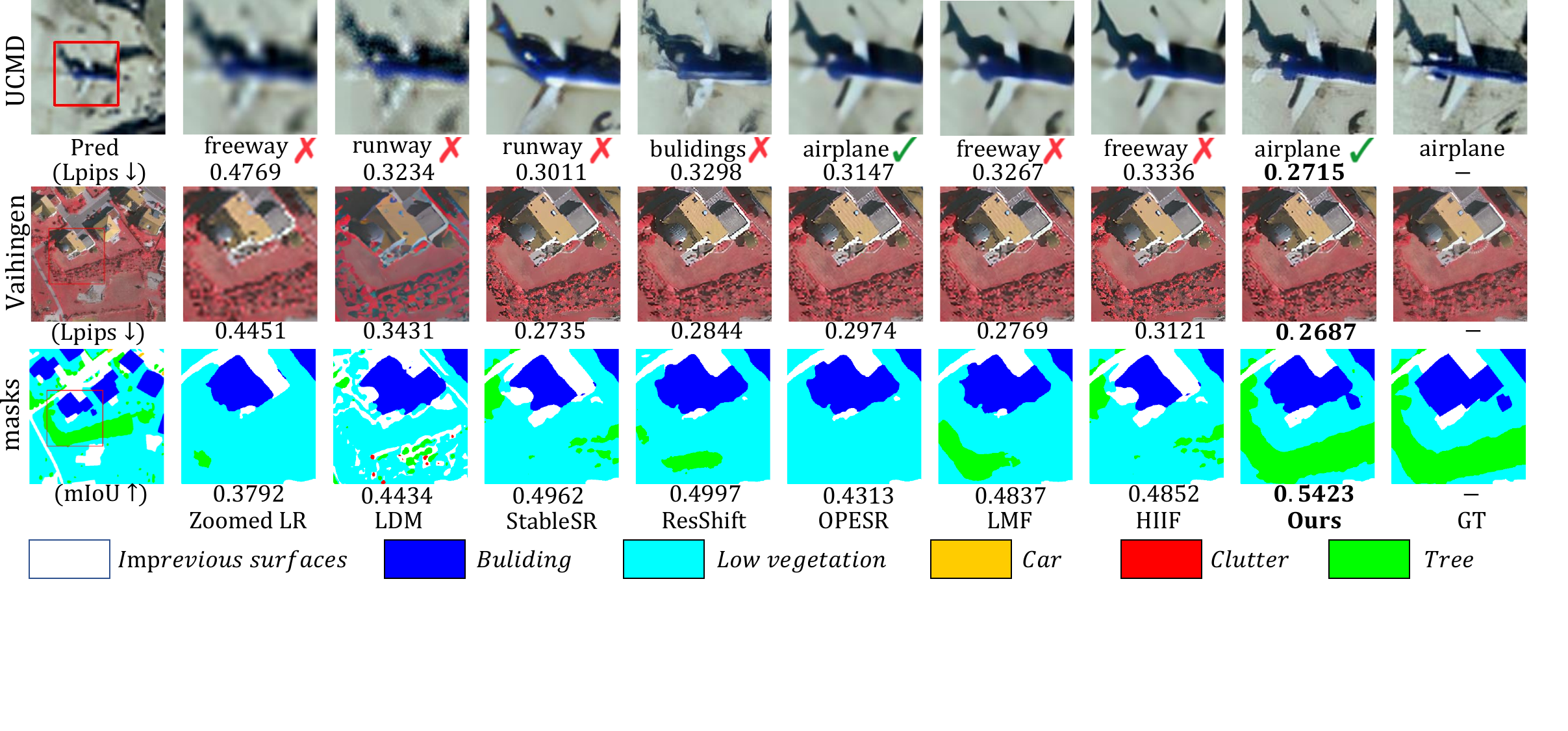}
    \caption{Qualitative comparisons of super-resolution and downstream task results under bicubic downsampling. From top to bottom, the rows correspond to spatial resolution enhancement from 4 m/px to 0.5 m/px ($\times 10$) and from 1.0 m/px to 0.1 m/px ($\times 5$), respectively.}
    \label{fig:seg}
\end{figure}


In this section, we provide the visual comparisons and their corresponding downstream task results under bicubic and real-world degradations in Fig.~\ref{fig:seg}. 
As shown in Fig.~\ref{fig:seg}, generative SR methods often struggle to preserve global structural integrity when the LR input contains limited reliable information, which in turn leads to hallucinated artifacts lacking physical consistency. 
Representative approaches such as StableSR and ResShift often compromise geometric structure in pursuit of texture realism, with higher upscaling scales inducing significant hallucinations and structural distortions.
These hallucinations can severely degrade downstream performance.

The structural fidelity of INR-based SR methods often comes at the cost of high-frequency detail, where oversmoothed textures lead to plastic reconstructions that impede remote sensing interpretation. 
Frameworks like LMF and HIIF exemplify this trade-off, producing outputs with insufficient richness to support the identification of small objects. 
The tree region in Fig.~\ref{fig:seg} illustrates this failure mode, where the lack of textural detail in the LMF reconstruction results in a segmentation miss.

On the contrary, the proposed SpatialSR generates high-quality images with faithful structure and realistic detail across nearly all SR scales and achieves superior performance at large magnification factors (e.g., ×10). Moreover, the results of SpatialSR show a strongly SRe-aware consistency, which strongly support the importance of SRe in downstream tasks. 

\subsection{Ablation Study}

\begin{table}[t]
\centering
\caption{Ablation study of SpatialSR on the LoveDA dataset. We separately apply the proposed SREM, RSCM, and TC loss to the LMF baseline, which serves as the shared backbone for fair comparison.}
\label{tab:ablation}
\setlength{\tabcolsep}{3pt}
\resizebox{\columnwidth}{!}{
\begin{tabular}{c|c|c|ccccc}
\toprule
\multirow{2}{*}{TSR} &
\multirow{2}{*}{CSR} &
\multirow{2}{*}{Method} &
\multicolumn{5}{c}{LoveDA} \\
\cmidrule(lr){4-8}
&
&
&
PSNR$\uparrow$ &
LPIPS$\downarrow$ &
NIQE$\downarrow$ &
MUSIQ$\uparrow$ &
mIoU$\uparrow$ \\
\midrule

\multirow{15}{*}{0.5}
& \multirow{5}{*}{2}
& baseline
& 29.89 & 0.3057 & 6.49 & 42.16 & 30.2 \\
&
& Ours + SREM
& 30.07 & 0.2953 & 6.20 & 44.63 & 31.5 \\
&
& Ours + RSCM
& 30.01 & \underline{0.2801} & 6.01 & 45.58 & \underline{32.5} \\
&
& Ours + TC
& \underline{30.19}
& 0.2869
& \underline{5.95}
& \underline{45.86}
& 32.9 \\
&
& Ours (SpatialSR)
& \textbf{30.25}
& \textbf{0.2798}
& \textbf{5.84}
& \textbf{46.08}
& \textbf{34.3} \\
\cmidrule{2-8}

& \multirow{5}{*}{4}
& baseline
& 24.87 & 0.5533 & 8.00 & 30.64 & 24.5 \\
&
& Ours + SREM
& 24.93 & \underline{0.5091} & 7.84 & 34.64 & 26.5 \\
&
& Ours + RSCM
& 25.10 & 0.5322 & 7.89 & \underline{35.99} & \underline{26.8} \\
&
& Ours + TC
& \underline{25.16}
& 0.5199
& \textbf{7.70}
& 35.58
& 27.7 \\
&
& Ours (SpatialSR)
& \textbf{25.20}
& \textbf{0.4947}
& \underline{7.73}
& \textbf{36.37}
& \textbf{28.4} \\
\cmidrule{2-8}

& \multirow{5}{*}{6}
& baseline
& 22.79 & 0.6836 & 9.64 & 24.24 & 19.4 \\
&
& Ours + SREM
& 23.00 & 0.5933 & 9.01 & 26.69 & 20.1 \\
&
& Ours + RSCM
& 23.01 & \textbf{0.5421} & 8.73 & 28.07 & 20.5 \\
&
& Ours + TC
& \textbf{23.10}
& 0.6299
& \underline{8.54}
& \underline{28.42}
& \underline{21.6} \\
&
& Ours (SpatialSR)
& \underline{23.06}
& \underline{0.5713}
& \textbf{8.48}
& \textbf{28.67}
& \textbf{21.9} \\
\bottomrule
\end{tabular}
}
\end{table}

In this section, we conduct a series of ablation experiments on the LoveDA-val datasets to dissect the individual contribution of each core component within SpatialSR. As illustrated in Table~\ref{tab:ablation}, we employ a component-wise incremental strategy to evaluate the effectiveness of the SREM, RSCM, and the TC loss.

\begin{table}[t]
\centering
\caption{Ablation study of SREM with different sampling strategies.}
\label{tab:ablation_sampling}
\setlength{\tabcolsep}{3pt}
\resizebox{\columnwidth}{!}{
\begin{tabular}{c|c|c|ccccc}
\toprule
\multirow{2}{*}{TSR} &
\multirow{2}{*}{CSR} &
\multirow{2}{*}{Method} &
\multicolumn{5}{c}{LoveDA} \\
\cmidrule(lr){4-8}
&
&
&
PSNR$\uparrow$ &
LPIPS$\downarrow$ &
NIQE$\downarrow$ &
MUSIQ$\uparrow$ &
mIoU$\uparrow$ \\
\midrule

\multirow{12}{*}{0.5}
& \multirow{4}{*}{2}
& SREM + Random sample
& 29.83 & 0.3156 & 6.32 & 42.89 & 30.7 \\
&
& SREM + Laplacian sample
& 29.90 & \underline{0.3020}
& 6.21
& 44.16
& 30.8 \\
&
& SREM + Haar sample
& \underline{30.01}
& 0.3087
& \textbf{6.18}
& \textbf{44.99}
& \underline{31.3} \\
&
& SREM + Hybrid sample (ours)
& \textbf{30.07}
& \textbf{0.2953}
& \underline{6.20}
& \underline{44.63}
& \textbf{31.5} \\
\cmidrule{2-8}

& \multirow{4}{*}{4}
& SREM + Random sample
& 24.76 & 0.5875 & 8.07 & 30.45 & 24.2 \\
&
& SREM + Laplacian sample
& 24.89 & 0.5423 & 8.00 & 31.99 & 25.1 \\
&
& SREM + Haar sample
& \textbf{24.95}
& \textbf{0.5034}
& \underline{7.92}
& \underline{33.81}
& \underline{25.8} \\
&
& SREM + Hybrid sample (ours)
& \underline{24.93}
& \underline{0.5091}
& \textbf{7.84}
& \textbf{34.64}
& \textbf{26.5} \\
\cmidrule{2-8}

& \multirow{4}{*}{6}
& SREM + Random sample
& 22.65 & 0.6892 & 9.52 & 24.36 & 19.4 \\
&
& SREM + Laplacian sample
& 22.87 & 0.6345 & \underline{9.22}
& 25.18
& \underline{19.7} \\
&
& SREM + Haar sample
& \underline{22.96}
& \underline{0.6099}
& 9.57
& \textbf{26.89}
& 19.6 \\
&
& SREM + Hybrid sample (ours)
& \textbf{23.00}
& \textbf{0.5933}
& \textbf{9.01}
& \underline{26.69}
& \textbf{20.1} \\
\bottomrule
\end{tabular}
}
\end{table}

\textit{1)} Effectiveness of SREM: To demonstrate the effectiveness of SREM, we first compare the performance of the proposed hybrid sampling strategy with individual random, Laplacian, and Haar sampling methods in Table~\ref{tab:ablation_sampling}. While global random sampling captures a spatially consistent representation of resolution-related cues, it remains insufficiently discriminative for fine-grained SRe estimation. 
Consequently, it yields only marginal improvements in no-reference IQA and mIoU compared to the baseline. As to the Laplacian and Haar sampling methods, they tend to identify high-frequency regions, which leads to significant improvements in fidelity (PSNR) and perceptual quality (LPIPS, NIQE, MUSIQ), especially at higher SR scales. However, Haar and Laplacian sampling primarily neglect the global spatial-resolution cues and therefore achieve unsatisfactory spatial-resolution estimation accuracy. Combining the advantages of the above methods, the hybrid strategy achieves the best performance on most of the metrics.

\begin{table}[t]
  \centering
  \caption{Ablation study of RSCM under different SRe style conditions.}
  \label{tab:RSCM_ablation}
  \resizebox{\columnwidth}{!}{
    \setlength{\tabcolsep}{4pt}
    \begin{tabular}{c|c|c|ccccc}
      \toprule
      \multirow{2}{*}{Target SRe} &
      \multirow{2}{*}{Current SRe} &
      \multirow{2}{*}{Method} &
      \multicolumn{5}{c}{LoveDa} \\
      \cmidrule(lr){4-8}
      & & &
      PSNR$\uparrow$ &
      LPIPS$\downarrow$ &
      NIQE$\downarrow$ &
      MUSIQ$\uparrow$ &
      mIoU$\uparrow$ \\
      \midrule

      \multirow{9}{*}{0.5}
      & \multirow{3}{*}{2}
      & RSCM (random SRe)
      & \underline{29.95}
      & \underline{0.3056}
      & 6.68
      & 42.89
      & 31.2 \\

      & & RSCM (w/o SRe)
      & 29.84
      & 0.3102
      & \underline{6.36}
      & \underline{44.21}
      & \underline{32.5} \\

      & & RSCM (correct SRe)
      & \textbf{30.01}
      & \textbf{0.2801}
      & \textbf{6.01}
      & \textbf{45.58}
      & \textbf{32.5} \\

      \cmidrule{2-8}

      & \multirow{3}{*}{4}
      & RSCM (random SRe)
      & \underline{24.95}
      & 0.5956
      & 8.10
      & 34.62
      & 26.7 \\

      & & RSCM (w/o SRe)
      & 24.82
      & \underline{0.5634}
      & \underline{7.99}
      & \underline{35.81}
      & \underline{27.1} \\

      & & RSCM (correct SRe)
      & \textbf{25.10}
      & \textbf{0.5322}
      & \textbf{7.89}
      & \textbf{35.99}
      & \textbf{26.8} \\

      \cmidrule{2-8}

      & \multirow{3}{*}{6}
      & RSCM (random SRe)
      & \underline{22.96}
      & 0.6453
      & 9.45
      & 27.92
      & 19.2 \\

      & & RSCM (w/o SRe)
      & 22.90
      & \underline{0.5756}
      & \underline{9.10}
      & \underline{27.99}
      & \underline{20.5} \\

      & & RSCM (correct SRe)
      & \textbf{23.01}
      & \textbf{0.5421}
      & \textbf{8.73}
      & \textbf{28.07}
      & \textbf{20.5} \\

      \bottomrule
    \end{tabular}
  }
\end{table}

\textit{2)}Ablation of RSCM: We investigate the effect of feature encoding under different styles, with results reported in Table \ref{tab:RSCM_ablation}. Compared with the baseline, modulating with random SRe style features yields a slight performance gain, because the random styles contain a small portion of features corresponding to the target SRe. The results indicate that using the correct target SRe style features yields the most pronounced improvements, with gains observed across all evaluation metrics.

\begin{table}[t]
  \centering
  \caption{Ablation study of the TC loss.}
  \label{tab:loss_ablation}
  \resizebox{\columnwidth}{!}{
    \setlength{\tabcolsep}{4pt}
    \begin{tabular}{c|c|c|ccccc}
      \toprule
      \multirow{2}{*}{Target SRe} &
      \multirow{2}{*}{Current SRe} &
      \multirow{2}{*}{Method} &
      \multicolumn{5}{c}{LoveDa} \\
      \cmidrule(lr){4-8}
      & & &
      PSNR$\uparrow$ &
      LPIPS$\downarrow$ &
      NIQE$\downarrow$ &
      MUSIQ$\uparrow$ &
      mIoU$\uparrow$ \\
      \midrule

      \multirow{9}{*}{0.5}
      & \multirow{3}{*}{2}
      & Pixel loss
      & \underline{30.09}
      & 0.3079
      & 6.69
      & 43.99
      & 30.6 \\

      & & Pixel + LPIPS
      & 29.82
      & \textbf{0.2799}
      & \underline{6.22}
      & \underline{44.01}
      & \underline{31.5} \\

      & & Pixel + TC (Ours)
      & \textbf{30.19}
      & \underline{0.2869}
      & \textbf{5.95}
      & \textbf{45.86}
      & \textbf{32.9} \\

      \cmidrule{2-8}

      & \multirow{3}{*}{4}
      & Pixel loss
      & \underline{25.10}
      & 0.5921
      & 7.99
      & 33.02
      & 24.9 \\

      & & Pixel + LPIPS
      & 24.82
      & \underline{0.5266}
      & \underline{7.85}
      & \underline{33.11}
      & \underline{26.1} \\

      & & Pixel + TC (Ours)
      & \textbf{25.16}
      & \textbf{0.5199}
      & \textbf{7.70}
      & \textbf{35.58}
      & \textbf{27.7} \\

      \cmidrule{2-8}

      & \multirow{3}{*}{6}
      & Pixel loss
      & \textbf{23.15}
      & 0.6933
      & 9.34
      & 24.24
      & 19.4 \\

      & & Pixel + LPIPS
      & 22.94
      & \textbf{0.6267}
      & \underline{9.00}
      & \underline{27.33}
      & \underline{20.5} \\

      & & Pixel + TC (Ours)
      & \underline{23.10}
      & \underline{0.6299}
      & \textbf{8.54}
      & \textbf{28.42}
      & \textbf{21.6} \\

      \bottomrule
    \end{tabular}
  }
\end{table}

\textit{3)}Ablation study of the TC loss. We investigate the effect of TC loss by replacing it with a standard perceptual loss. As shown in Table~\ref{tab:loss_ablation}, although perceptual loss improves visual quality, it introduces noticeable fidelity degradation. 
specifically, on the LoveDA dataset, it reduces NIQE by 0.80 under the challenging SRe adaptation setting from 6\,m/px to 0.5\,m/px (CSR\,=\,6\,m/px, i.e., $\times 12$ SR), clearly demonstrating the importance of the proposed TC loss for preserving realistic texture details. Furthermore, as detailed in the supplementary material (Table~\ref{tab:cal_complexity}), SpatialSR achieves a single-step inference time of 0.0074\,s with 1.76\,M parameters and 51.96\,G FLOPs on $128\!\rightarrow\!512$ SR, remaining competitive with the lightest INR-based method LMF (0.0062\,s, 1.39\,M parameters) while delivering substantial improvements in SR quality and downstream performance.

%% file: sec/5_conclusion.tex
\section{Conclusion and Limitations}
In this paper, we propose SpatialSR, a SRe-centric RSI-ASSR pipeline. In SpatialSR, we first estimate the absolute SRe of the input LR-RSI via a Spatial Resolution Estimation Module, and further compute the scale factor according to the predicted SRe and target SRe. Next, we develop the RSCM module to reconstruct the SR-RSI guided by the resolution-aware latent variables extracted from a pretrained SRe-to-image diffusion model. Finally, we propose the trajectory consistency loss to leverage the strong generative capability and texture synthesis behavior of diffusion models, which introduces more realistic high-frequency details without iterative sampling. Extensive experiments demonstrate the universal effectiveness and efficiency of our SpatialSR on SR benchmarks and downstream tasks such as segmentation and classification. Furthermore, we built a real-world RSI-ASSR dataset named MSRUAV to verify the generalization of our method.

Although the proposed SpatialSR method achieves state-of-the-art results on most of the SR and downstream metrics, a limitation remains. Our approach struggles to handle target SRe values beyond the predefined resolution range, which limits the practical applicability of SREM and RSCM. This issue becomes more pronounced in cross-sensor scenarios, where estimation accuracy degrades due to distribution shifts. A potential direction is to model how image details evolve across continuous SRe, which may improve generalization beyond the predefined range. 


%% file: sec/6_sup.tex


\section{The Complete Derivation for TC Loss}

Given the clean latent variable $X_{0}$, its noisy counterpart $\hat{X}_{0}$, and Gaussian noise $\epsilon \sim \mathcal{N}(0, 1)$, our goal is to formulate a TC loss that bridges the generative trajectory of real data $X_T$ and its predicted counterpart $\hat{X}_T$. We aim to model the conditional consistency between the two posterior transitions:
\begin{equation}
\mathcal{L} = \mathcal{L}\left( P(X_{T-1}|X_{T}),\, P(\hat{X}_{T-1}|\hat{X}_{T}) \right).
\end{equation}
However, the two posterior distributions are not directly comparable due to their distinct conditioning variables. To address this, we reformulate the objective as
\begin{equation}
\mathcal{L} = \mathcal{L}\left( P(X_{T-1}|X_{T},\hat{X}_{0}),\, P(\hat{X}_{T-1}|\hat{X}_{T},\hat{X}_{0}) \right),
\label{eq:main_objective_app}
\end{equation}
which allows us to condition both paths on $\hat{X}_0$ for consistent probabilistic comparison.

\noindent According to the DDPM formulation, the model-generated posterior can be expressed as:
\begin{equation}
\begin{aligned}
P(\hat{X}_{T-1}|\hat{X}_{T},\hat{X}_{0})
=
\mathcal{N}\Bigg(
&
\frac{1}{\sqrt{\alpha_{T}}}
\left(
\hat{X}_{T}
-\frac{1-\alpha_{T}}
{\sqrt{1-\overline{\alpha}_{T}}}
\epsilon
\right),
\\
&
\frac{(1-\alpha_{T})
(1-\overline{\alpha}_{T-1})}
{1-\overline{\alpha}_{T}}
\Bigg).
\end{aligned}
\end{equation}

\noindent To obtain a comparable true path, we compute $P(X_{T-1}|X_{T},\hat{X}_{0})$ using Bayes’ theorem:
\begin{equation}
P(X_{T-1}|X_{T},\hat{X}_{0}) 
= \frac{P(X_{T}|X_{T-1},\hat{X}_{0})\,P(X_{T-1}|\hat{X}_{0})}{P(X_{T}|\hat{X}_{0})}.
\end{equation}
Since the forward diffusion process is Markovian, we have
\begin{equation}
P(X_{T}|X_{T-1},\hat{X}_{0}) = P(X_{T}|X_{T-1}) = \mathcal{N}\left(\sqrt{\alpha_{T}}X_{T-1},\, 1-\alpha_{T}\right).
\end{equation}
The remaining two terms are marginal likelihoods, obtained by integrating out $X_0$:
\begin{align}
P(X_{T-1}|\hat{X}_{0}) &= \int P(X_{T-1}|X_{0})\, P(X_{0}|\hat{X}_{0})\, dX_{0}, \label{eq:marginal1}\\
P(X_{T}|\hat{X}_{0}) &= \int P(X_{T}|X_{0})\, P(X_{0}|\hat{X}_{0})\, dX_{0}. \label{eq:marginal2}
\end{align}
From the forward process of DDPM, we know:
\begin{align}
P(X_{T-1}|X_{0}) &= \mathcal{N}\left(\sqrt{\overline{\alpha}_{T-1}}X_{0},\, 1-\overline{\alpha}_{T-1}\right), \\
P(X_{T}|X_{0}) &= \mathcal{N}\left(\sqrt{\overline{\alpha}_{T}}X_{0},\, 1-\overline{\alpha}_{T}\right).
\end{align}

\noindent To solve the integrals in Eqs.~\ref{eq:marginal1}--\ref{eq:marginal2}, we first determine $P(X_{0}|\hat{X}_{0})$. Using Bayes’ rule,
\begin{equation}
P(X_{0}|\hat{X}_{0}) = \frac{P(\hat{X}_{0}|X_{0})P(X_{0})}{P(\hat{X}_{0})}.
\end{equation}
We adopt a tractable local residual model that treats $\hat{X}_{0}$ as a noisy observation of $X_{0}$, defined as $\hat{X}_{0}=X_{0}+\epsilon'$ with $\epsilon' \sim \mathcal{N}(0, {\sigma'}^{2})$. This Gaussian residual form is a modeling assumption rather than a derived consequence of the observation model, adopted for tractability. Therefore,
\begin{equation}
P(\hat{X}_{0}|X_{0}) = \mathcal{N}(X_{0}, {\sigma'}^{2}).
\end{equation}
Combined with a Gaussian prior $P(X_{0}) \sim \mathcal{N}(0, I)$, the posterior $P(X_{0}|\hat{X}_{0})$ remains Gaussian and, for simplicity, is approximated as
\begin{equation}
P(X_{0}|\hat{X}_{0}) \approx \mathcal{N}(\hat{X}_{0}, {\sigma'}^{2}).
\end{equation}

\noindent Given this approximation, the marginal likelihoods become Gaussian through convolution:
\begin{align}
P(X_{T-1}|\hat{X}_{0}) &= \mathcal{N}\!\left(\sqrt{\overline{\alpha}_{T-1}}\hat{X}_{0},\, 1-\overline{\alpha}_{T-1}+\overline{\alpha}_{T-1}{\sigma'}^{2}\right), \\
P(X_{T}|\hat{X}_{0}) &= \mathcal{N}\!\left(\sqrt{\overline{\alpha}_{T}}\hat{X}_{0},\, 1-\overline{\alpha}_{T}+\overline{\alpha}_{T}{\sigma'}^{2}\right).
\end{align}

\noindent Substituting the above into the Bayesian posterior gives:
\begin{equation}
\begin{split}
P(X_{T-1}|X_{T},\hat{X}_{0})
=
\frac{
\mathcal{N}
\left(
\sqrt{\alpha_{T}}X_{T-1},
1-\alpha_{T}
\right)
}{
\mathcal{N}
\left(
\sqrt{\overline{\alpha}_{T}}\hat{X}_{0},
1-\overline{\alpha}_{T}
+\overline{\alpha}_{T}{\sigma'}^{2}
\right)
}
\\
\times
\mathcal{N}
\left(
\sqrt{\overline{\alpha}_{T-1}}\hat{X}_{0},
1-\overline{\alpha}_{T-1}
+\overline{\alpha}_{T-1}{\sigma'}^{2}
\right).
\end{split}
\end{equation}

\noindent This posterior is proportional to the product of two Gaussian densities, which yields another Gaussian:
\begin{align}
P(X_{T-1}|X_{T},\hat{X}_{0})
\propto
\exp\Bigg(
-\frac{1}{2}
\Bigg[
&
\frac{
(X_{T}-\sqrt{\alpha_{T}}X_{T-1})^{2}
}{
1-\alpha_{T}
}
\nonumber\\
&+
\frac{
(X_{T-1}-\sqrt{\overline{\alpha}_{T-1}}\hat{X}_{0})^{2}
}{
1-\overline{\alpha}_{T-1}
+\overline{\alpha}_{T-1}{\sigma'}^{2}
}
\Bigg]
\Bigg).
\end{align}
By completing the square in $X_{T-1}$ and comparing to the standard Gaussian form, we identify the variance and mean as:
\begin{align}
\sigma^{2} &= \frac{(1-\alpha_{T})(1-\overline{\alpha}_{T-1}+\overline{\alpha}_{T-1}{\sigma'}^{2})}
{1-\overline{\alpha}_{T}+\overline{\alpha}_{T}{\sigma'}^{2}}, \\
\mu &= 
\frac{\sqrt{\overline{\alpha}_{T-1}}(1-\alpha_{T})}{1-\overline{\alpha}_{T}+\overline{\alpha}_{T}{\sigma'}^{2}}\hat{X}_{0}
+ \frac{(1-\overline{\alpha}_{T-1}+\overline{\alpha}_{T-1}{\sigma'}^{2})\sqrt{\alpha_{T}}}
{1-\overline{\alpha}_{T}+\overline{\alpha}_{T}{\sigma'}^{2}}X_{T}.
\label{eq:posterior_mean_app}
\end{align}

\noindent Next, substituting $\hat{X}_{0} = X_{0} + \epsilon' = \frac{X_{T}-\sqrt{1-\overline{\alpha}_{T}}\epsilon}{\sqrt{\overline{\alpha}_{T}}} + \epsilon'$ into Eq.~\ref{eq:posterior_mean_app} yields:
\begin{equation}
\mu = \left(\frac{A}{\sqrt{\overline{\alpha}_{T}}} + B\right)X_{T}
- \frac{A\sqrt{1-\overline{\alpha}_{T}}}{\sqrt{\overline{\alpha}_{T}}}\epsilon + A\epsilon',
\end{equation}
where
\begin{align}
A &= \frac{\sqrt{\overline{\alpha}_{T-1}}(1-\alpha_{T})}{1-\overline{\alpha}_{T}+\overline{\alpha}_{T}{\sigma'}^{2}}, \\
B &= \frac{(1-\overline{\alpha}_{T-1}+\overline{\alpha}_{T-1}{\sigma'}^{2})\sqrt{\alpha_{T}}}{1-\overline{\alpha}_{T}+\overline{\alpha}_{T}{\sigma'}^{2}}.
\end{align}
Here, $\epsilon$ denotes the forward diffusion noise, while $\epsilon'$ represents the modeling error in $\hat{X}_{0}$.

\noindent Finally, the posterior distribution can be summarized as:
\begin{equation}
P(X_{T-1}|X_{T},\hat{X}_{0}) = \mathcal{N}\!\left( \mu,\, \sigma^{2} \right).
\end{equation}
\noindent\textbf{Justification of mean matching.}
The TC loss aligns the two posterior distributions through their means, which raises the question of whether the variance/covariance discrepancy is silently ignored. We now show that this term is provably inert. For two isotropic Gaussians $\mathcal{N}(\mu_1,\sigma_1^2 I)$ and $\mathcal{N}(\mu_2,\sigma_2^2 I)$ in $\mathbb{R}^{d}$, the KL divergence decomposes as
\begin{multline}
D_{\mathrm{KL}}\!\left(\mathcal{N}(\mu_1,\sigma_1^2 I)\,\|\,\mathcal{N}(\mu_2,\sigma_2^2 I)\right) \\
= \frac{1}{2}\,d\Big(\tfrac{\sigma_1^2}{\sigma_2^2}-1-\ln\tfrac{\sigma_1^2}{\sigma_2^2}\Big) \\
+ \frac{1}{2\sigma_2^2}\lVert\mu_1-\mu_2\rVert^{2}.
\end{multline}
The first term is the variance/covariance contribution: it is non-negative and vanishes if and only if $\sigma_1^2=\sigma_2^2$. In our formulation, the model posterior adopts the DDPM fixed reverse variance $\beta_{T}=\frac{(1-\alpha_{T})(1-\overline{\alpha}_{T-1})}{1-\overline{\alpha}_{T}}$, whereas the true posterior variance is $\sigma^{2}=\frac{(1-\alpha_{T})(1-\overline{\alpha}_{T-1}+\overline{\alpha}_{T-1}{\sigma'}^{2})}{1-\overline{\alpha}_{T}+\overline{\alpha}_{T}{\sigma'}^{2}}$. Two facts follow.
(i) As ${\sigma'}^{2}\to0$, we have $\sigma^{2}\to\beta_{T}$, so the variance term vanishes exactly and the KL reduces to mean matching.
(ii) Even when ${\sigma'}^{2}\neq0$, both $\sigma^{2}$ and $\beta_{T}$ depend only on the noise schedule $\{\alpha_{T},\overline{\alpha}_{T},\overline{\alpha}_{T-1}\}$ and the fixed residual variance ${\sigma'}^{2}$, none of which depends on the learnable network parameters $\theta$; the variance term is therefore a constant with respect to $\theta$. Denoting the posterior means of the true and predicted distributions by $\mu$ and $\hat{\mu}$, respectively, we therefore have
\begin{equation}
\begin{split}
&\arg\min_{\theta}\,D_{\mathrm{KL}}\!\left(P(X_{T-1}|X_{T},\hat{X}_{0})\right. \\
&\quad \left.\,\|\,P(\hat{X}_{T-1}|\hat{X}_{T},\hat{X}_{0})\right) = \arg\min_{\theta}\,\lVert\mu-\hat{\mu}\rVert^{2}.
\end{split}
\end{equation}
Hence, minimizing the mean-squared error between the posterior means is not a lossy simplification that discards the covariance; it is the exact parameter-dependent component of the distributional objective, and it is consistent with the DDPM simplified objective, which likewise drops the constant variance weighting.

Consequently, the Trajectory-Consistency (TC) loss $L_{\text{TC}}$ is defined as the mean-squared error between the posterior means of the true and predicted distributions. Denoting the mean of the model posterior by $\hat{\mu}$, i.e.,
\begin{equation}
\hat{\mu}
=
\frac{\sqrt{\overline{\alpha}_{T-1}}(1-\alpha_{T})}{1-\overline{\alpha}_{T}}\hat{X}_{0}
+
\frac{(1-\overline{\alpha}_{T-1})\sqrt{\alpha_{T}}}{1-\overline{\alpha}_{T}}\hat{X}_{T},
\end{equation}
and the mean of the true posterior by $\mu$ (Eq.~\eqref{eq:posterior_mean_app}), the TC loss reduces to
\begin{equation}
L_{\text{TC}}
=
\lVert \mu - \hat{\mu} \rVert^{2}.
\end{equation}
This formulation provides a unified probabilistic interpretation of the TC loss, ensuring consistency between the real and predicted diffusion trajectories. Additionally, our total loss incorporates pixel-level losses and utilizes the parameter $\lambda$ to ensure that TC Loss and pixel loss are of the same magnitude, achieving a more balanced perception-distortion trade-off.

\begin{equation}
\mathcal{L} = \mathcal{L}_{\text{pixel}} + \lambda \cdot \mathcal{L}_{\text{TC}}
\label{eq:total_loss_app}
\end{equation}

\section{Ablation on the Perception-Distortion Trade-off}
We study how the weight $\lambda$ of the trajectory-consistency loss balances pixel fidelity and perceptual quality. Following the perceptual metrics used throughout this paper, we define a composite Perceptual Score that unifies the full-reference and no-reference indicators into a single higher-is-better quantity:
\begin{equation}
\begin{aligned}
\mathrm{Score} =& \ (1 - \mathrm{LPIPS}) + (1 - \mathrm{DISTS}) + \mathrm{CLIP\text{-}IQA} \\
&+ \mathrm{MANIQA} + \frac{\mathrm{MUSIQ}}{100} + \max\left(0, \frac{10 - \mathrm{NIQE}}{10}\right),
\end{aligned}
\label{eq:score}
\end{equation}
where the full-reference terms LPIPS and DISTS are subtracted from $1$, MUSIQ is rescaled to $[0,1]$, and NIQE is clipped so that a lower NIQE yields a higher score.

\begin{table}[t]
\centering
\caption{Ablation study on the TC loss weight $\lambda$. Evaluated on UCMD for 4$\times$ SR (1.2 m/px $\rightarrow$ 0.3 m/px).}
\label{tab:lambda_ablation}
\begin{tabular}{c|ccc}
\toprule
$\lambda$ & PSNR$\uparrow$ & SSIM$\uparrow$ & Score$\uparrow$ \\
\midrule
0    & 28.9803 & 0.7842 & 2.6752  \\
0.05 & 28.9543 & 0.7759 & 2.6651  \\
0.1  & 28.9217 & 0.7683 & 2.8914  \\
0.2  & 28.8542 & 0.7628 & 2.9286  \\
0.3  & 28.8358 & 0.7616 & 3.0205  \\
0.4  & 28.7858 & 0.7582 & 3.0857  \\
0.5  & 28.6894 & 0.7551 & 3.0929  \\
\bottomrule
\end{tabular}
\end{table}

Based on the ablation results presented in Table~\ref{tab:lambda_ablation}, we observe a clear perception-distortion trade-off as the weight $\lambda$ of the trajectory-consistency loss increases. With higher $\lambda$ values, the perceptual quality improves consistently, as reflected by the progressive increase in the Perceptual Score. Conversely, pixel-level fidelity metrics exhibit a corresponding decline. The configuration with $\lambda = 0.3$ achieves an optimal balance between perceptual enhancement and distortion control, delivering competitive perceptual quality while maintaining reasonable fidelity. Therefore, we select $\lambda = 0.3$ as the default setting in our final model, striking the best trade-off for downstream applications.

\section{Additional Experimental Details}
\label{app:exp_details}
\begin{figure}[t]
    \centering
    \includegraphics[trim=5px 20px 10px 0, clip,width=\linewidth]{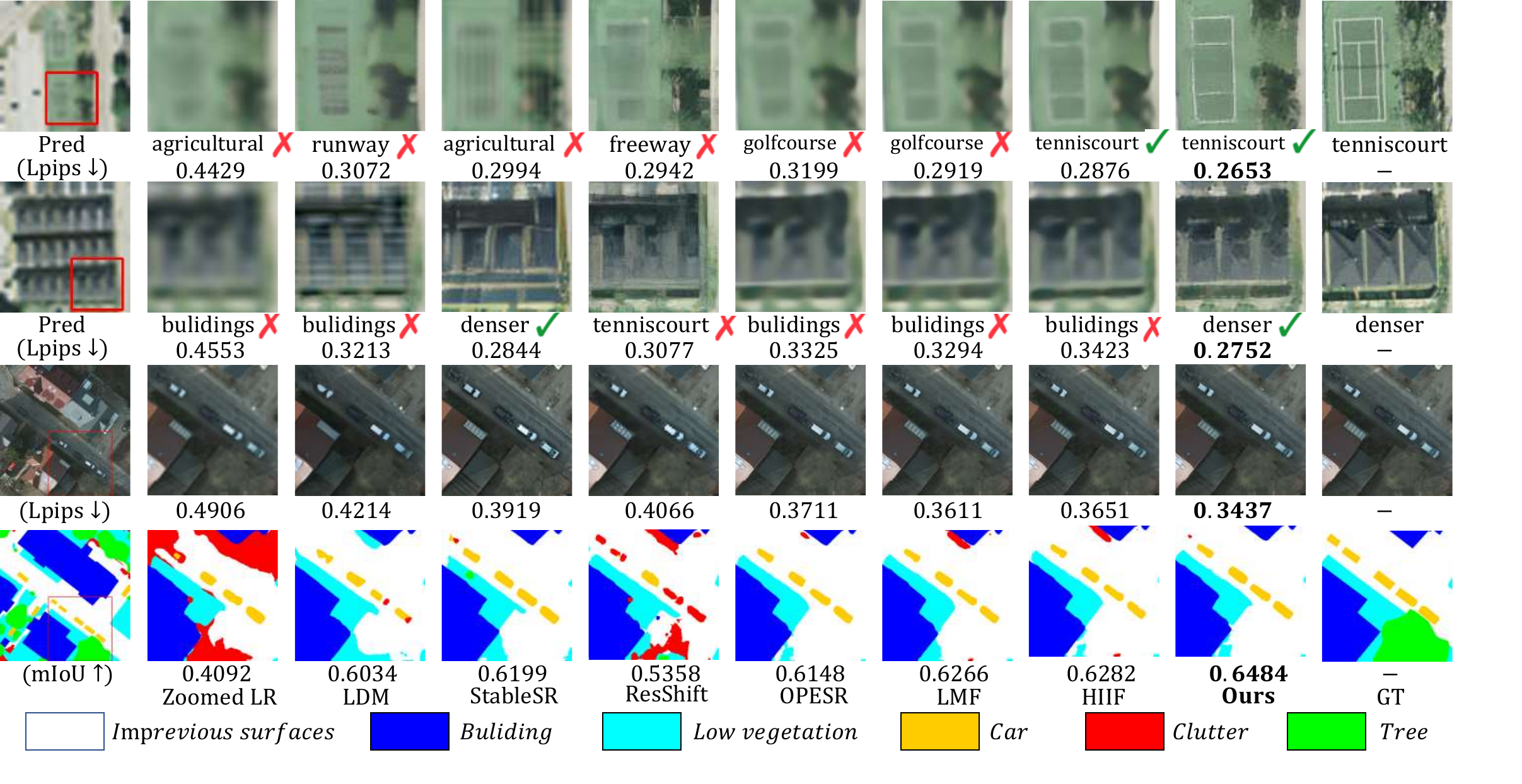}
    \caption{Qualitative comparisons of super-resolution and downstream task results under bicubic downsampling. From top to bottom, the rows correspond to spatial resolution enhancement from 2.0 m/px to 0.5 m/px ($\times 4$) ,from 4.0 m/px to 0.5 m/px ($\times 8$) and from 0.5 m/px to 0.1 m/px ($\times 5$), respectively.}
    \label{fig:seg_sup}
\end{figure}

\textit{1)} Datasets: In addition to the public benchmarks used for evaluation, we introduce a Multi-Spatial Resolution Unmanned Aerial Vehicle (MSRUAV) dataset to support model training and further validate the robustness of the proposed method. MSRUAV consists of 3,036 real-world UAV images and 4,096 samples collected from Google Earth, covering spatial resolutions ranging from 0.1 to 2.0 m/pixel. The dataset is divided into training, validation, and test sets with a ratio of 7:1:2.

\textit{2)} Implementation Details:
To simulate realistic imaging conditions, arbitrary-scale LR-HR pairs are synthesized using the Real-ESRGAN degradation pipeline and bicubic downsampling. 
During training, the degradation scale is uniformly sampled from a continuous range of $[1\times,20\times]$. HR images are randomly cropped and downsampled to generate 48$\times$48 LR patches. 
All models are implemented in PyTorch and trained for 1,000 epochs on two NVIDIA RTX 4090 GPUs with a batch size of 128. The learning rate is initialized at 1e-4 and multiplied by 0.5 every 200 epochs. To provide resolution-specific priors for the Resolution-aware Style Control Module (RSCM), we pre-extract resolution-aware latent codes from the proposed SRe-to-image diffusion model. The extracted latent features are organized into an offline lookup dictionary spanning spatial resolutions from 0.1 to 2.0 m/px, where each resolution level contains 1,000 style embeddings. The SRe-to-image diffusion model is built upon Stable Diffusion v1.5 and fine-tuned using the training split of the MSRUAV dataset.



\section{Complete Experimental Results}

\begin{table}[t]
\centering
\caption{Quantitative comparison results of SR and segmentation performance on bicubic-degraded synthetic datasets. The best and second-best results are shown in bold and underlined, respectively.}
\setlength{\tabcolsep}{2.2pt}
\renewcommand{\arraystretch}{0.9}
\small
\resizebox{\textwidth}{!}{\begin{tabular}{c|c|cc|ccccc|ccccccccc}
\toprule
\multirow{2}{*}{Dataset} & \multirow{2}{*}{TSR} & \multirow{2}{*}{CSR} & \multirow{2}{*}{Metric}
& \multicolumn{5}{c}{Generative SR}
& \multicolumn{8}{c}{Implicit Neural Representation-based SR} \\
\cmidrule(r){5-9} \cmidrule(l){10-17}
& & & & ESRGAN & IDM & LDM & StableSR & ResShift & LIIF & ContSR & AnySR & SRNO & LMF & HIIF & Ours & LR \\
\midrule
\multirow{24}{*}{Potsdam}
& \multirow{24}{*}{0.1}
& \multirow{6}{*}{0.5}
& PSNR↑ & 26.91 & 26.16 & 28.80 & 28.14 & 30.22 & 32.69 & 32.32 & 32.71 & 31.70 & \underline{33.05} & 32.45 & \textbf{33.22} & & \\
& & & SSIM↑ & 0.6185 & 0.6160 & 0.7735 & 0.7652 & 0.8065 & 0.8767 & 0.8742 & 0.8756 & 0.8678 & \underline{0.8779} & 0.8746 & \textbf{0.8955}& &  \\
          &       &       & LPIPS↓ & 0.3926 & 0.4965 & 0.3752 & 0.3012 & \textbf{0.2436} & 0.2734 & 0.2732 & 0.2734 & 0.2751 & 0.2726 & 0.2677 & \underline{0.2591} &       &  \\
          &       &       & DISTS↓ & 0.1474 & 0.2514 & 0.1929 & 0.1526 & 0.1557 & 0.1256 & 0.1261 & \underline{0.1222} & 0.1285 & 0.1276 & 0.1239 & \textbf{0.1183} &       &  \\
& & & FID↓ & 10.64 & 47.85 & 37.09 & 32.06 & 17.97 & 4.19 & 4.32 & \underline{4.15} & 5.40 & 4.25 & 4.39 & \textbf{3.74} & & \\
& & & mIoU↑ & 71.3 & 68.1 & 70.4 & 73.4 & 74.9 & 75.8 & 76.3 & 76.4 & 72.3 & \underline{76.5} & 75.2 & \textbf{80.2} & 72.7 \\
\cmidrule{3-17}
& & \multirow{6}{*}{1}
& PSNR↑ & 25.71 & 25.40 & 27.09 & 27.32 & 27.18 & 29.71 & 29.68 & 30.07 & 29.26 & \underline{30.15} & 29.79 & \textbf{30.30} & & \\
          &       &       & SSIM↑ & 0.5070 & 0.6087 & 0.7407 & 0.7138 & 0.7118 & 0.7993 & 0.7998 & 0.8011 & 0.7945 & \underline{0.8015} & 0.8006 & \textbf{0.8331} &       &  \\
          &       &       & LPIPS↓ & 0.4458 & 0.5889 & 0.4996 & \textbf{0.4072} & 0.4225 & 0.4507 & 0.4491 & 0.4568 & 0.4366 & 0.4691 & 0.4375 & \underline{0.4173} &       &  \\
          &       &       & DISTS↓ & 0.2121 & 0.2937 & 0.2617 & \textbf{0.1916} & 0.2247 & 0.2268 & 0.2301 & 0.2245 & 0.2285 & 0.2368 & 0.2253 & \underline{0.2100} &       &  \\
& & & FID↓ & 39.57 & 69.84 & 58.56 & 43.40 & 43.81 & 39.86 & 40.69 & 40.13 & 41.81 & 41.49 & \underline{39.53} & \textbf{39.42} & & \\
& & & mIoU↑ & 42.8 & 33.5 & 38.9 & \underline{49.7} & 49.4 & 47.2 & 45.7 & 49.3 & 41.2 & 48.6 & 45.6 & \textbf{55.3} & 35.3 \\
\cmidrule{3-17}
& & \multirow{6}{*}{1.5}
& PSNR↑ & 23.61 & 23.36 & 23.50 & 24.55 & 26.76 & 28.07 & 28.17 & 28.29 & 27.88 & \underline{28.30} & 28.16 & \textbf{28.54} & & \\
          &       &       & SSIM↑ & 0.6423 & 0.6087 & 0.7087 & 0.6977 & 0.6900  & 0.7599 & 0.7638 & 0.7637 & 0.7605 & \underline{0.7640} & 0.7636 & \textbf{0.7924} &       &  \\
          &       &       & LPIPS↓ & 0.6065 & 0.6961 & \textbf{0.4999} & 0.5427 & 0.5509 & 0.5873 & 0.5946 & 0.6122 & 0.5787 & 0.6231 & 0.5809 & \underline{0.5259} &       &  \\
          &       &       & DISTS↓ & 0.3463 & 0.3768 & \textbf{0.2579} & 0.3008 & \underline{0.2831} & 0.3348 & 0.3375 & 0.3333 & 0.3281 & 0.3505 & 0.3270 & 0.3083 &       &  \\
& & & FID↓ & 88.50 & 104.52 & \underline{88.18} & 96.22 & 90.71 & 106.45 & 114.47 & 102.09 & 116.66 & 107.37 & 111.04 & \textbf{84.93} & & \\
& & & mIoU↑ & 27.2 & 27.4 & 31.1 & \underline{31.8} & 30.3 & 23.6 & 23.4 & 26.3 & 26.3 & 21.3 & 25.5 & \textbf{33.4} & 15.2 \\
\cmidrule{3-17}
& & \multirow{6}{*}{2}
& PSNR↑ & 23.43 & 23.11 & 23.44 & 24.37 & 25.97 & 27.18 & 27.16 & \underline{27.44} & 26.91 & 27.43 & 27.09 & \textbf{27.78} & & \\
          &       &       & SSIM↑ & 0.6474 & 0.5578 & 0.6992 & 0.7050 & 0.6728 & 0.7526 & 0.7511 & 0.7528 & 0.7475 & \underline{0.7533} & 0.7493 & \textbf{0.7959} &       &  \\
          &       &       & LPIPS↓ & 0.7135 & 0.6024 & \underline{0.5470} & 0.6055 & 0.5970 & 0.6329 & 0.6434 & 0.6617 & 0.6264 & 0.6681 & 0.6257 & \textbf{0.5364} &       &  \\
          &       &       & DISTS↓ & 0.3567 & 0.3188 & \textbf{0.2912} & \underline{0.3001} & 0.3045 & 0.3849 & 0.3858 & 0.3853 & 0.3712 & 0.4031 & 0.3686 & 0.3438 &       &  \\
& & & FID↓ & 141.54 & 123.25 & \underline{109.37} & 111.88 & 121.69 & 144.19 & 153.82 & 136.13 & 155.45 & 146.17 & 147.70 & \textbf{108.26} & & \\
& & & mIoU↑ & 16.8 & 15.2 & 17.1 & \underline{21.5} & 20.1 & 16.2 & 15.9 & 18.1 & 20.8 & 15.4 & 18.6 & \textbf{26.7} & 5.1 \\
\midrule
\multirow{24}{*}{LoveDa}
& \multirow{24}{*}{0.5}
& \multirow{6}{*}{2}
& PSNR↑ & 27.61 & 27.09 & 27.31 & 28.97 & 29.84 & 29.92 & 29.46 & \underline{30.06} & 29.60 & 29.89 & 29.70 & \textbf{30.25} &  \\
          &       &       & SSIM↑ & 0.7331 & 0.6871 & 0.7031 & 0.7559 & 0.7703 & 0.8137 & 0.8119 & \underline{0.8155} & 0.7976 & 0.8143 & 0.8103 & \textbf{0.8231} &  \\
          &       &       & LPIPS↓ & 0.5807 & 0.5055 & 0.3945 & 0.3861 & 0.3244 & 0.2937 & 0.2898 & \underline{0.2823} & 0.3015 & 0.3057 & 0.2855 & \textbf{0.2798} &  \\
          &       &       & DISTS↓ & 0.2674 & 0.2444 & 0.1908 & 0.2108 & 0.1765 & 0.1673 & 0.1659 & 0.1597 & 0.1647 & 0.1752 & \underline{0.1573} & \textbf{0.1554} &  \\
          &       &       & FID↓  & 59.05 & 84.42 & 32.92 & 34.50 & 21.12 & 17.68 & 17.10 & 17.24 & 23.77 & 17.49 & \underline{17.01} & \textbf{15.57} &  \\
          &       &       & mIoU↑ & 28.5  & 29.3  & 29.8  & 32.4  & \underline{33.2} & 29.4  & 28.8  & 29.7  & 31.2  & 30.2  & 31.8  & \textbf{34.3} & 21.7 \\
\cmidrule{3-17}
& & \multirow{6}{*}{4}
& PSNR↑ & \underline{25.19} & 25.05 & 25.13 & 25.17 & 25.18 & 24.71 & 24.66 & 24.59 & 24.93 & 24.87 & 24.84 & \textbf{25.20} &  \\
          &       &       & SSIM↑ & \underline{0.6918} & 0.6033 & 0.6629 & 0.6099 & 0.6504 & 0.6698 & 0.6604 & 0.6777 & 0.6687 & 0.6761 & 0.6699 & \textbf{0.7032} &  \\
          &       &       & LPIPS↓ & 0.6327 & 0.6045 & \underline{0.4922} & 0.5001 & \textbf{0.4815} & 0.5327 & 0.5303 & 0.5465 & 0.5015 & 0.5533 & 0.5049 & 0.4947 &  \\
          &       &       & DISTS↓ & 0.3587 & 0.3237 & \textbf{0.2065} & 0.2475 & 0.2334 & 0.2766 & 0.2716 & 0.2723 & 0.2665 & 0.2792 & 0.2678 & \underline{0.2245} &  \\
          &       &       & FID↓  & 97.58 & 84.75 & 52.89 & 60.38 & \textbf{48.07} & 70.35 & 71.18 & 69.35 & 71.01 & 72.72 & 72.62 & \underline{51.21} &  \\
          &       &       & mIoU↑ & 20.1  & 21.1  & 26.5  & \underline{27.8} & 27.6  & 24.9  & 22.8  & 25.6  & 25.0    & 24.5  & 23.1  & \textbf{28.4} & 11.7 \\
\cmidrule{3-17}
& & \multirow{6}{*}{6}
& PSNR↑ & 21.58 & 21.68 & 22.56 & 22.04 & \underline{22.94} & 22.29 & 21.77 & 22.74 & 22.13 & 22.79 & 21.83 & \textbf{23.06} &  \\
          &       &       & SSIM↑ & 0.6135 & 0.5969 & 0.6348 & 0.5834 & 0.6128 & 0.6357 & 0.6199 & \underline{0.6415} & 0.6278 & 0.6414 & 0.6252 & \textbf{0.6713} &  \\
          &       &       & LPIPS↓ & 0.6962 & 0.6116 & \textbf{0.5451} & \underline{0.5534} & 0.5970 & 0.6600  & 0.6702 & 0.6884 & 0.6581 & 0.6836 & 0.6349 & 0.5713 &  \\
          &       &       & DISTS↓ & 0.4249 & 0.3823 & 0.2964 & \textbf{0.2739} & 0.2857 & 0.3309 & 0.3237 & 0.3279 & 0.3153 & 0.3326 & 0.3147 & \underline{0.2847} &  \\
          &       &       & FID↓  & 130.12 & 121.75 & \textbf{96.98} & 98.52 & 101.70 & 120.45 & 132.78 & 117.01 & 117.55 & 125.34 & 117.98 & \underline{98.25} &  \\
          &       &       & mIoU↑ & 19.5  & 17.9  & 19.6  & \underline{21.2} & 19.5  & 18.4  & 18.9  & 19.5  & 19.3  & 19.4  & 19.0    & \textbf{21.9} & 7.3 \\
\cmidrule{3-17}
& & \multirow{6}{*}{8}
& PSNR↑ & 20.81 & 20.66 & 22.46 & \underline{22.48} & 21.91 & 21.24 & 20.75 & 21.70 & 21.11 & 21.77 & 20.75 & \textbf{22.52} &  \\
          &       &       & SSIM↑ & 0.5257 & 0.5365 & 0.6118 & 0.5041 & 0.5961 & 0.6218 & 0.6068 & 0.6281 & 0.6137 & \underline{0.6294} & 0.6106  & \textbf{0.6681} &  \\
          &       &       & LPIPS↓ & 0.7741 & 0.7619 & 0.7066 & 0.7055 & \textbf{0.6906} & 0.7278 & 0.7398 & 0.7611 & 0.7438 & 0.7506 & 0.7214 & \underline{0.6972} &  \\
          &       &       & DISTS↓ & 0.3871 & 0.4123 & 0.4469 & 0.4483 & \underline{0.3451} & 0.3668 & 0.3618 & 0.3675 & 0.3522 & 0.3699 & 0.3488 & \textbf{0.3447} &  \\
          &       &       & FID↓  & 168.72 & 173.31 & \textbf{159.49} & 166.69 & 165.80 & 173.59 & 179.32 & 165.36 & 165.55 & 176.52 & 165.62 & \underline{164.92} &  \\
          &       &       & mIoU↑ & 4.0     & 4.8   & 9.2   & \underline{9.3} & 8.1     & 7.1   & 7.7   & 7.8   & 7.9   & 6.9   & 7.9   & \textbf{9.5} & 3.7 \\
\bottomrule
\end{tabular}}
\label{tab:complete}
\end{table}
We further report quantitative results under target spatial resolutions of 0.1 m/px and 0.5 m/px on the validation sets of Potsdam and LoveDA, respectively (Table~\ref{tab:complete}). On Potsdam, our method surpasses the current state-of-the-art StableSR by 2.98 dB PSNR in the ×10 SR setting, demonstrating substantially improved reconstruction fidelity. For downstream scene understanding, our approach also achieves higher mIoU on Potsdam, outperforming the previous best method (StableSR) by 11.2\% in the ×10 SR setting. All downstream models are applied to the SR images in a zero-shot manner, directly using their officially released weights without any additional fine-tuning. These results confirm the effectiveness of the proposed SREM for target spatial-resolution-aware reconstruction and its benefits for downstream tasks.

To complement the quantitative results, Fig.~\ref{fig:seg_sup} presents visual comparisons of different methods. Consistent with Table~\ref{tab:complete}, SpatialSR produces high-quality reconstructions with more faithful structures and realistic textures across different SR scales. The superiority is particularly evident in challenging large-magnification settings (e.g., ×10), where competing methods often exhibit artifacts or structural distortions. Furthermore, SpatialSR maintains strong target-SRe consistency, highlighting the effectiveness of incorporating spatial-resolution awareness into the SR process.

\section{Complexity Analysis}

\begin{table}[t]
  \centering
  \caption{\textbf{Complexity comparison among different methods.} ContinuousSR is abbreviated as ContSR. For our method, the reported inference time includes the sliding-window cropping and patch-selection steps of SREM.}
  \small
  \setlength{\tabcolsep}{2pt}
  \resizebox{\textwidth}{!}{
  \begin{tabular}{c|cccccccccccc}
    \toprule
    $128 \rightarrow 512$ & ESRGAN & IDM   & LDM   & StableSR & ResShift & LIIF  & ContSR   & AnySR & SRNO  & LMF & HIIF  & Ours \\
    \midrule
    Trainable Params(M) & 16.7  & 116.6 & 123.6 & 150   & 40.6  & \underline{1.57} & 1.71  & 6.43  & 2.02  & \textbf{1.39} & 2.96  & 1.76 \\
    Inference Time(S) & 0.033 & 35.87 & 27.92 & 25.91 & 6.67  & 0.1084 & 0.1921 & 0.0638 & 0.056 & \textbf{0.0062} & 0.2041 & \underline{0.0074} \\
    MACs(G) & 293.71 & 3920.41 & 3918.92 & 4087.52 & 454.65 & 382.62 & 153.2 & 105.42 & 231.3 & \textbf{24.37} & 440.1 & \underline{25.76} \\
    FLOPs(G) & 587.87 & 7944.61 & 7936.11 & 8175.33 & 808.82 & 766.63 & 307.72 & 211.58 & 463.9 & \textbf{49.16} & 885.5 & \underline{51.96} \\
    Inference Step & /     & 150   & 150   & 150   & 4     & /     & /     & /     & /     & /     & /     & / \\
    \bottomrule
  \end{tabular}}
  \label{tab:cal_complexity}
\end{table}

Table \ref{tab:cal_complexity} compares the computational complexity of SpatialSR and competing methods. We report the inference time for 4× SR on 128×128 LR inputs using a single NVIDIA RTX 4090 GPU, with 150 sampling steps for diffusion-based methods. Due to the iterative denoising process and heavy pretrained backbones, diffusion-based methods exhibit higher inference costs and parameter counts. In contrast, the INR-based method LMF achieves the lowest complexity and fastest inference. Benefiting from the lightweight design of SREM and RSCM, SpatialSR attains the second-lowest inference time and computational complexity while delivering substantial improvements in both SR quality and downstream performance.

\section{Generalization and Sampling Strategy Analysis of \mbox{SREM}}
\begin{figure}[!t]
    \centering
    \includegraphics[trim=0px 0px 0px 0, clip,width=\columnwidth]{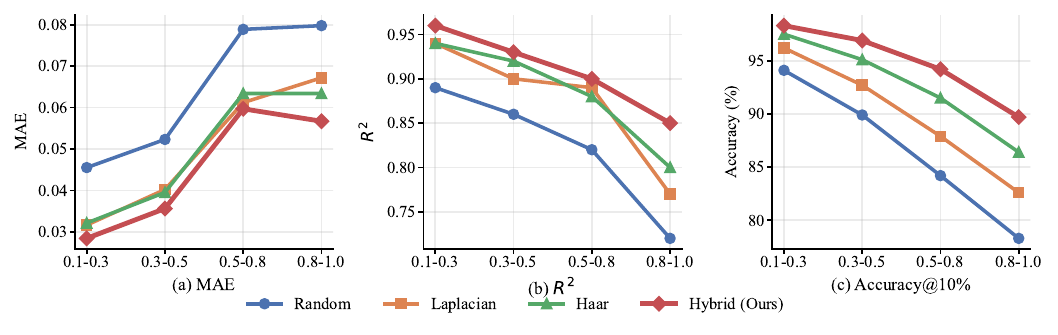}
    \caption{Ablation study of different sampling strategies under varying spatial-resolution ranges. (a) MAE, (b) $R^2$, and (c) Accuracy within ±10\%. The horizontal axis denotes the spatial-resolution range (m/px). Hybrid sampling consistently achieves the best performance across all evaluation metrics.}
    \label{fig:sampling_strategy_tmm}
\end{figure}

To verify the effectiveness of the selected hybrid sampling strategy for estimating the spatial resolution degradation of LR images, we freeze the weights of SRENet and adopt different sampling strategies for its input, respectively, and evaluate their estimation accuracy on the MSRUAV validation set. 
As shown in Fig~\ref{fig:sampling_strategy_tmm}, although both Haar Sample and Laplacian Sample can analyze the high-frequency information of images and achieve relatively good prediction accuracy, they mainly focus on local high-frequency components while overlooking the global spatial-resolution cues, thereby resulting in unsatisfactory spatial-resolution estimation accuracy. By combining the advantages of the above methods, the hybrid strategy achieves the best performance on most metrics.

\begin{figure}[!t]
    \centering
    \includegraphics[trim=10px 75px 75px 0, clip,width=\columnwidth]{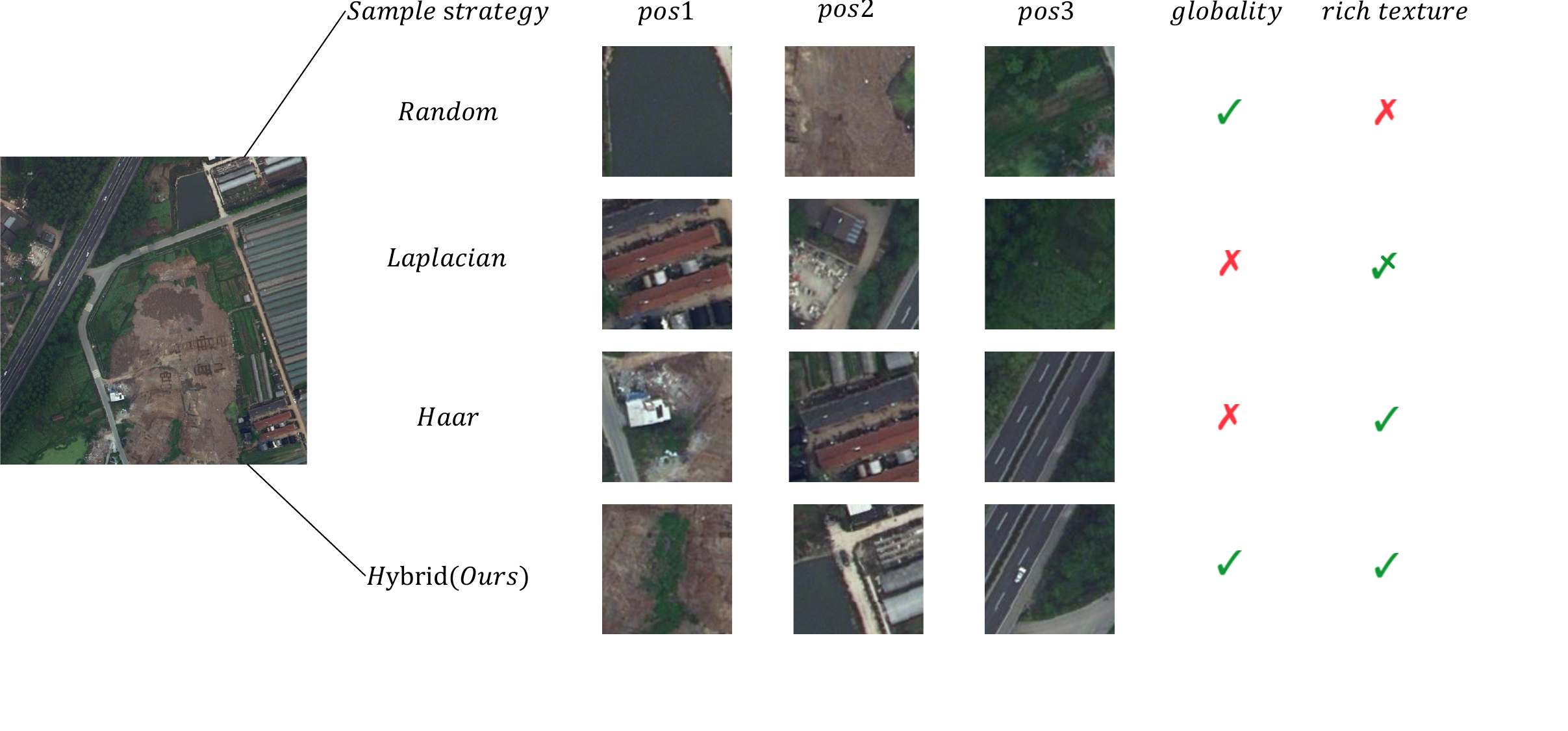}
    \caption{Visualizations of sampling results under different sampling strategies in the SREM for SRe estimation. The Random strategy produces samples dominated by low-frequency information, thereby increasing the difficulty of resolution estimation. Although the Laplacian and Haar strategies capture more high-frequency details, they fail to preserve the global characteristics of spatial resolution, leading to reduced estimation accuracy.}
    \label{fig:sample}
\end{figure}

We further provide visualizations of the samples collected by the four different sampling strategies, as shown in Fig.~\ref{fig:sample}. Our hybrid sample not only preserves the high-frequency details of the LR image, but also maintains the consistency of the global structural cues of the LR image.

\begin{table}[t]
\centering
\caption{Cross-scenario and cross-domain generalization of SREM. Within-domain settings serve as upper-bound references.}
\resizebox{\columnwidth}{!}{
\begin{tabular}{llccccc}
\toprule
\multirow{2}{*}{Train} & \multirow{2}{*}{Test} &
\multirow{2}{*}{MAE$\downarrow$} & \multirow{2}{*}{RMSE$\downarrow$} &
\multirow{2}{*}{$R^2\uparrow$} & \multicolumn{2}{c}{Accuracy$\uparrow$} \\
\cmidrule(lr){6-7}
& & & & & $\pm$5\% & $\pm$10\% \\
\midrule
MSRUAV (UAV)   & MSRUAV (UAV)     & 0.0312 & 0.0401 & 0.94 & 87.6 & 96.8 \\
MSRUAV (UAV)   & UCMD (satellite) & 0.0487 & 0.0623 & 0.88 & 74.3 & 89.5 \\
UCMD-urban     & UCMD-urban       & 0.0298 & 0.0387 & 0.95 & 89.1 & 97.2 \\
UCMD-urban     & UCMD-forest      & 0.0441 & 0.0573 & 0.90 & 79.8 & 92.4 \\
\bottomrule
\end{tabular}}
\label{tab:srem_generalization}
\end{table}

To further investigate whether SREM learns a genuine physical SRe prior rather than a dataset-specific statistic, we evaluate its cross-scenario and cross-domain generalization, as reported in Table~\ref{tab:srem_generalization}. The within-domain setting (MSRUAV$\rightarrow$MSRUAV) attains the highest estimation accuracy (MAE\,=\,0.0312, $R^2$\,=\,0.94) and serves as an upper-bound reference. When transferred across domains from UAV to satellite imagery (MSRUAV$\rightarrow$UCMD), the estimation accuracy decreases to MAE\,=\,0.0487 and $R^2$\,=\,0.88, indicating a moderate drop that nevertheless remains well above a trivial estimator, suggesting that SREM captures resolution-related frequency cues that transfer across sensor types. 
Similarly, the cross-scenario transfer within UCMD from urban to forest scenes (UCMD-urban$\rightarrow$UCMD-forest) yields MAE\,=\,0.0441 and $R^2$\,=\,0.90, suggesting that the model relies on resolution-correlated frequency statistics rather than scene-specific textures. These results indicate that SREM relies on resolution-related frequency priors rather than scene-specific textures.

\begin{figure}[t]
    \centering
    \includegraphics[trim=8px 10px 10px 15px, clip,width=\columnwidth]{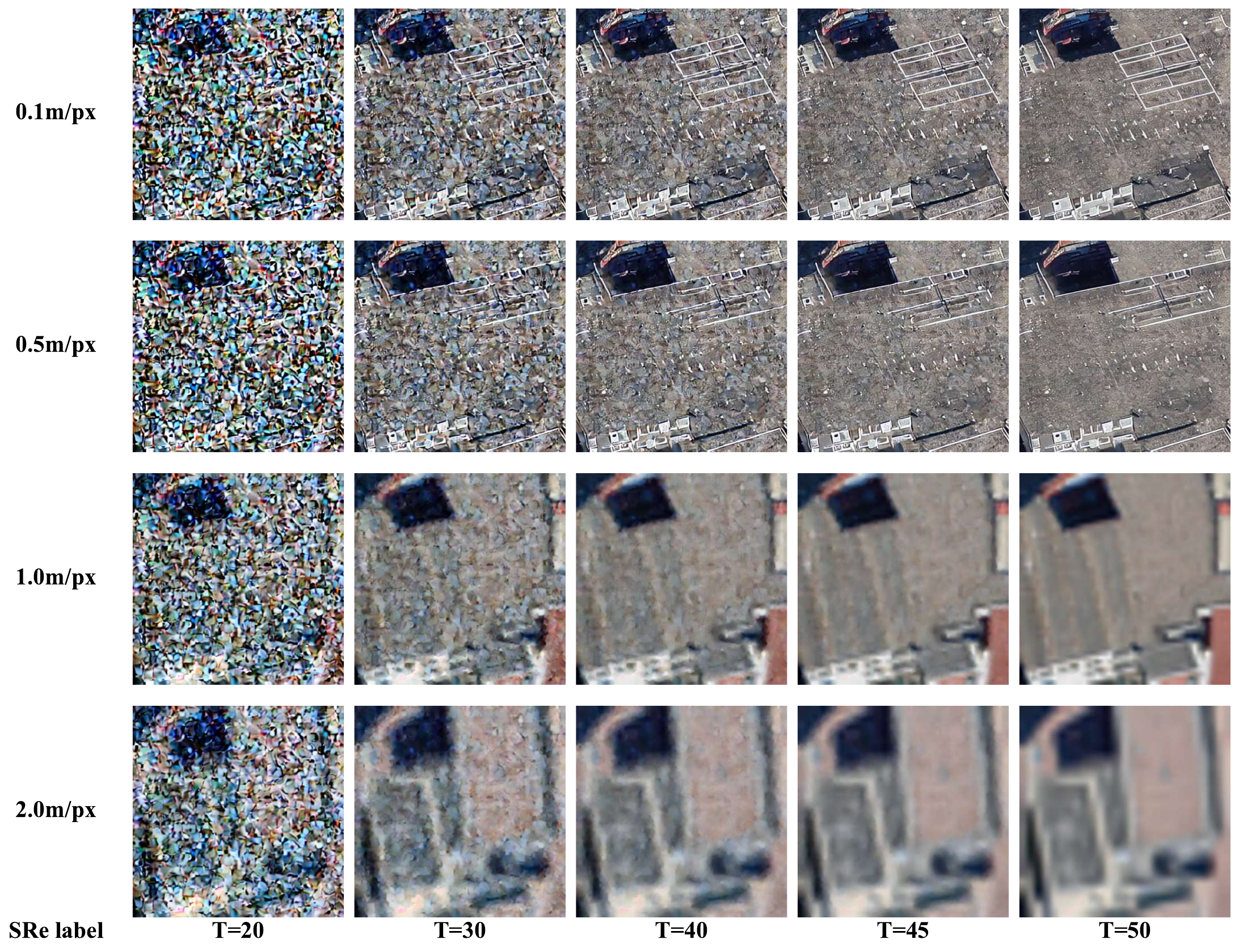}
    \caption{Visualizations of the same scene across continuous spatial resolutions and their corresponding diffusion timesteps $T$. The visualization at $T=45$ preserves both texture details and weak semantic structures, making its latent code the most suitable for style modulation.}
    \label{fig:sre-to-img}
\end{figure}

\section{Ablation Study of the Diffusion Prior in RSCM}
To govern the citicality of resolution-aware latent codes, Fig.~\ref{fig:sre-to-img} visualizes the same scene under different SRe conditions, validating our training procedure and supporting our feature selection. Specifically, we visualize the latent features at timesteps 20, 30, 40, 45, and 50 under different prompts, and evaluate them from the perspectives of texture richness and semantic complexity. For the super-resolution task, the style modulation encoding is expected to preserve rich texture information while maintaining relatively weak semantic content for effective modulation. From the visualization results, the latent features at (T=20, 30,) and (40) lack sufficient texture details, whereas the latent feature at (T=50) contains excessively strong semantic information. Therefore, the latent feature extracted at (T=45) is considered the most suitable choice.

\begin{table}[!t]
\centering
\caption{Ablation study on the diffusion prior in RSCM. ``GT SRe'' uses the ground-truth spatial resolution; ``est.\ SRe'' uses the SREM prediction.}
\resizebox{\columnwidth}{!}{
\begin{tabular}{lccccc}
\toprule
Variant & PSNR$\uparrow$ & SSIM$\uparrow$ & LPIPS$\downarrow$ & DISTS$\downarrow$ & FID$\downarrow$ \\
\midrule
(a) Pretrained CNN feature  & 32.81 & 0.885 & 0.271 & 0.126 & 4.18 \\
(a) CLIP feature            & 32.74 & 0.882 & 0.274 & 0.128 & 4.23 \\
(b) Misaligned style        & 32.69 & 0.879 & 0.283 & 0.133 & 4.41 \\
(c) w/o SRe fine-tuning     & 32.77 & 0.883 & 0.275 & 0.129 & 4.31 \\
(d) GT SRe + diffusion      & 33.29 & 0.896 & 0.256 & 0.117 & 3.70 \\
\textbf{Ours (est.\ SRe + diffusion)} & \textbf{33.22} & \textbf{0.895} & \textbf{0.259} & \textbf{0.118} & \textbf{3.74} \\
\bottomrule
\end{tabular}}
\label{tab:rscm_ablation}
\end{table}

To disentangle the source of RSCM's gain, we conduct four ablations on the diffusion prior (Table~\ref{tab:rscm_ablation}). (a) Replacing the diffusion feature with a pretrained CNN or CLIP feature degrades PSNR from 33.22 to 32.81/32.74\,dB, showing that the improvement is not merely due to a strong pretrained feature. (b) Guiding the modulation with a misaligned style embedding reduces PSNR to 32.69\,dB and raises LPIPS to 0.283, confirming that correctly matched SRe style codes are essential. (c) Using a diffusion feature without SRe fine-tuning yields 32.77\,dB, validating the effectiveness of the SRe conditioning step. (d) Replacing the estimated SRe with the ground-truth SRe yields only a marginal change of 0.07\,dB in PSNR (33.22 vs.\ 33.29), indicating that the SREM estimation error has a limited impact on the final SR quality. These results collectively demonstrate that the RSCM gain stems from the SRe-specific diffusion prior rather than an arbitrary strong feature.